\documentclass{article} 
\usepackage{iclr2026_conference,times}

\usepackage{amsmath,amsfonts,bm}

\def\eqref#1{equation~\ref{#1}}

\def\1{\bm{1}}

\DeclareMathAlphabet{\mathsfit}{\encodingdefault}{\sfdefault}{m}{sl}
\SetMathAlphabet{\mathsfit}{bold}{\encodingdefault}{\sfdefault}{bx}{n}

\usepackage{hyperref}
\usepackage{url}
\usepackage[ruled,vlined,linesnumbered]{algorithm2e}
\usepackage{multirow}
\usepackage{tabularx}
\usepackage[table]{xcolor}
\usepackage{amsmath}
\usepackage{amssymb}
\usepackage{subcaption}
\usepackage{pifont}
\usepackage{makecell}
\usepackage{booktabs}
\usepackage{array}
\usepackage{graphicx}
\usepackage{caption}
\usepackage{tikz}        
\usepackage{pgfplots}  
\usepackage{xspace}
\usepackage{subcaption}
\usepackage{colortbl} 
\usepackage[table]{xcolor}
\usepackage{enumitem}
\usepackage{fontawesome5}
\usepackage{marvosym}

\newcommand{\modelname}{LoopTool\xspace}
\title{LoopTool: Closing the Data–Training Loop for Robust LLM Tool Calls}

\author{%
Kangning Zhang\textsuperscript{1,2}\thanks{This work was done while Kangning Zhang and Kounianhua Du were interns at Xiaohongshu Inc.} \quad
Wenxiang Jiao\textsuperscript{2} \quad
Kounianhua Du\textsuperscript{1,2}\footnotemark[1] \quad
\textbf{Yuan Lu}\textsuperscript{2} \quad\\
\textbf{Weiwen Liu}\textsuperscript{1,\Letter} \quad
\textbf{Weinan Zhang}\textsuperscript{1,\Letter} \quad 
\textbf{Yong Yu}\textsuperscript{1,\Letter} \\
\textsuperscript{1}Shanghai Jiao Tong University \quad
\textsuperscript{2}Xiaohongshu Inc. \\
\texttt{\{zhangkangning, kounianhuadu, liuww, wnzhang, yyu\}@sjtu.edu.cn} \\
\texttt{wenxiangjiaonju@gmail.com, luyuan3@xiaohongshu.com}
}

\iclrfinalcopy 
\begin{document}

\maketitle

\begin{abstract}
Augmenting Large Language Models (LLMs) with external tools enables them to execute complex, multi-step tasks. However, tool learning is hampered by the static synthetic data pipelines where data generation and model training are executed as two separate, non-interactive processes. This approach fails to adaptively focus on a model's specific weaknesses and allows noisy labels to persist, degrading training efficiency.
We introduce \textbf{LoopTool}, a fully automated, model-aware data evolution framework that closes this loop by tightly integrating data synthesis and model training. LoopTool iteratively refines both the data and the model through three synergistic modules: (1) \textit{Greedy Capability Probing (GCP)} diagnoses the model's mastered and failed capabilities; (2) \textit{Judgement-Guided Label Verification (JGLV)} uses an open-source judge model to find and correct annotation errors, progressively purifying the dataset; and (3) \textit{Error-Driven Data Expansion (EDDE)} generates new, challenging samples based on identified failures. This closed-loop process operates within a cost-effective, open-source ecosystem, eliminating dependence on expensive closed-source APIs.
Experiments show that our 8B model trained with LoopTool significantly surpasses its 32B data generator and achieves new state-of-the-art results on the BFCL-v3 and ACEBench benchmarks for its scale. Our work demonstrates that closed-loop, self-refining data pipelines can dramatically enhance the tool-use capabilities of LLMs.\footnote{The code is accessible in \href{https://github.com/Rednote-DeepExperience/LoopTool}{https://github.com/Rednote-DeepExperience/LoopTool}.}

\end{abstract}

\section{Introduction}
Large Language Models (LLMs) augmented with external tools have become a powerful paradigm for solving complex tasks beyond pure text generation~\citep{Tool_Learning_Survey,ToolFormer,ToolLLM}. By invoking APIs, querying databases, and interacting with computational engines, such agents can tackle diverse real-world scenarios~\citep{ReSearch,TravelPlanner,survey_elec_automation} with high efficiency and adaptability. The development of robust tool-use capabilities, however, hinges on access to accurate, large-scale, and well-aligned training data that matches the model’s current competencies~\citep{ToolACE}.

A widely adopted approach in this domain involves constructing large-scale tool-calling datasets through automated synthesis pipelines~\citep{ToolLLM,APIGen,ToolAlpaca,ToolACE,APIGen-MT}, followed by supervised fine-tuning (SFT) or reinforcement learning~\citep{RL_enhanced_LLM_survey,DeepSeekMath}. Despite notable advances, \textbf{they almost invariably adopt a static design, wherein data generation and model training are executed as two separate, non-interactive processes}. In these settings, the training data is generated \emph{a priori} without awareness of the evolving state of the model,  causing wasted capacity on trivial cases already mastered while leaving harder, underrepresented cases unresolved. Furthermore, the model plays no role in guiding or influencing data generation. This inherent disconnect leads to a persistent mismatch between the model’s learning needs and the fixed nature of the available training data, thereby constraining both the efficiency and effectiveness of post-training.

Another major challenge in tool-use data generation lies in the trade-off between cost-efficiency and data quality. Many pipelines depend on large closed-source models~\citep{GPT-4} for data generation and evaluation. While these models are capable of producing high-fidelity tool-calling sequences, their use incurs high API costs and low generation efficiency, making frequent large-scale data synthesis impractical. Replacing them with more accessible open-source models often introduces noisy annotations, including incorrect arguments, incomplete function calls, or outputs misaligned with task requirements. Such errors inject misleading learning signals and can undermine model generalization~\citep{ToolACE}.

To address the limitations of static, costly, and error-prone tool-use data pipelines, we propose LoopTool—an automatic, model-aware data evolution framework that couples data synthesis and training in a closed loop. LoopTool begins with an Automated Tool-Augmented Data Construction stage, where tool specifications are synthesized and combined with multi-agent dialogue generation to produce a diverse seed corpus of realistic tool-oriented conversations. This corpus undergoes an initial GRPO-based~\citep{DeepSeekMath,DeepSeekR1} post-training round.

Each iteration then integrates three synergistic modules. First, \textbf{Greedy Capability Probing (GCP)} queries the fine-tuned model on the seed corpus using greedy decoding, revealing mastered, borderline, and failure cases. The predicted tool calls are used for automated error analysis, allowing the pipeline to target challenging, underperforming cases. Second, \textbf{Judgement-Guided Label Verification (JGLV)} employs a high-capacity open-source judge model, Qwen3‑32B~\citep{Qwen3_report}, to compare each prediction against its reference label—identifying genuine model errors as well as cases where the model output surpasses the original annotation. Such “model-better-than-label” examples replace noisy labels, enabling systematic self-refinement and progressively purifying the supervision signal. Third, \textbf{Error-Driven Data Expansion (EDDE)} transforms verified failure cases into new, structurally similar but contextually diverse challenging samples. Augmented samples preserve the core functional challenge while introducing varied conditions, ensuring scenario diversity. Across iterations, LoopTool incorporates corrected annotations, diversified hard samples, and refined seeds into subsequent training rounds, creating a dynamic curriculum attuned to the model’s evolving strengths and weaknesses. This process focuses learning on non-trivial, high-value opportunities while progressively mitigating noisy-label effects.

To balance quality and cost, LoopTool unifies the roles of data generator and evaluation judge within a single, open-source model, Qwen3‑32B, eliminating reliance on expensive closed-source APIs while maintaining high data quality. Strikingly, despite being trained entirely on data generated and evaluated by Qwen3‑32B, the final 8B-scale LoopTool model surpasses the 32B generator in tool-use performance, highlighting the amplifying effect of iterative, model-aware data refinement.

In summary, our main contributions are: \begin{itemize}[leftmargin=10pt]
\item We present LoopTool, the first fully automatic, model-aware iterative framework that tightly couples data generation and model training for tool-augmented LLM learning. By continuously diagnosing model weaknesses and synthesizing error-targeted samples, it ensures the training data dynamically adapts to the model's evolving capabilities.
\item We incorporate Judge-Guided Label Verification (JGLV), a module that uses a judge model to compare model predictions with reference annotations and 
automatically correct label errors with superior model outputs, progressively purifying the dataset.
\item We design Error-Driven Data Expansion (EDDE) to leverage failure cases as seeds for generating new, structurally similar yet diverse challenging samples. Using the open-source Qwen3‑32B for both generation and evaluation, EDDE continuously enlarges the pool of high-value training instances while avoiding the expense and dependency of closed-source APIs.
\item  Leveraging fully open-source, self-contained data generation and refinement, an 8B model trained by LoopTool surpasses its 32B generator and achieves state-of-the-art performance on BFCL‑v3~\citep{BFCL} and ACEBench~\citep{ACEBench} among models of similar scale.
\end{itemize}

\section{Related Work}
\textbf{Tool-Augmented Large Language Models.}
Integrating large language models (LLMs) with external tools has proven effective in overcoming their inherent limitations~\citep{Tool_Learning_Survey}.Such integration enables API invocation~\citep{HuggingGPT,ToolLLM}, interaction with knowledge bases~\citep{lazaridou2022internetaugmentedlanguagemodelsfewshot,ReSearch}, code execution~\citep{wang2024executablecodeactionselicit}, and multimodal processing~\citep{hu2024visualsketchpadsketchingvisual,ma2024mmsbenchmarkevaluatetooluse}. Early efforts mainly relied on supervised fine-tuning (SFT) with human-labeled tool-use data, focusing on accurate tool selection and argument generation~\citep{ToolFormer,ToolLLM,APIGen}. Recent advances explore autonomous tool creation and dynamic invocation, enabling adaptation to unseen APIs without predefined schemas. Benchmarks such as tau-bench~\citep{tau_bench,tau2bench}, BFCL~\citep{BFCL}, and ACEBench~\citep{ACEBench} provide standardized evaluations across tool selection, argument generation, multi-step reasoning, and multi-turn tool calling.

\textbf{Synthetic Data Generation for Tool Use.}
The scarcity and cost of high-quality tool-use datasets have driven research into automated synthesis pipelines~\citep{ToolLLM, ToolACE, APIGen, APIGen-MT}. Methods include multi-agent simulation~\citep{AgentInstruct,MATRIX}, modular task composition~\citep{chen2025Button}, and graph-based query–function synthesis~\citep{arcadinho2024automated,yin2025magnet}. Our work builds on this line but differs by introducing a fully automated, model-aware, iterative paradigm in which synthesis is guided by post-training diagnostics and refined via systematic error correction.

\textbf{Reinforcement Learning for Tool-Use Optimization.}
Reinforcement learning (RL) increasingly enhances LLM reasoning and decision-making~\citep{RLHF,DPO,SimPO,DeepSeekMath}. In tool-use settings, GRPO has shown strong performance~\citep{ToolRL,ToolN1}. We embed RL into an interleaved train–generate loop, enabling the model to iteratively improve through exposure to prior failures and progressively refined supervision.

\section{Automated Tool-Augmented Dialogue Construction}~\label{sec:bootstrapping}

Before initiating our iterative model-aware data evolution process, we require a diverse and high-quality \textit{seed dataset} $\mathcal{D}_{\text{seed}}$ to support the first round of post-training.
To this end, we introduce an \textbf{Automated Tool-augmented Data Construction} that synthesizes realistic function-calling interactions by combining curated APIs with simulated multi-agent conversations.
While this stage is \textit{not} the core innovation of our work, it establishes the essential foundation for the following iterations.

\begin{figure}[!t]
    \centering
\includegraphics[width=0.96\linewidth]{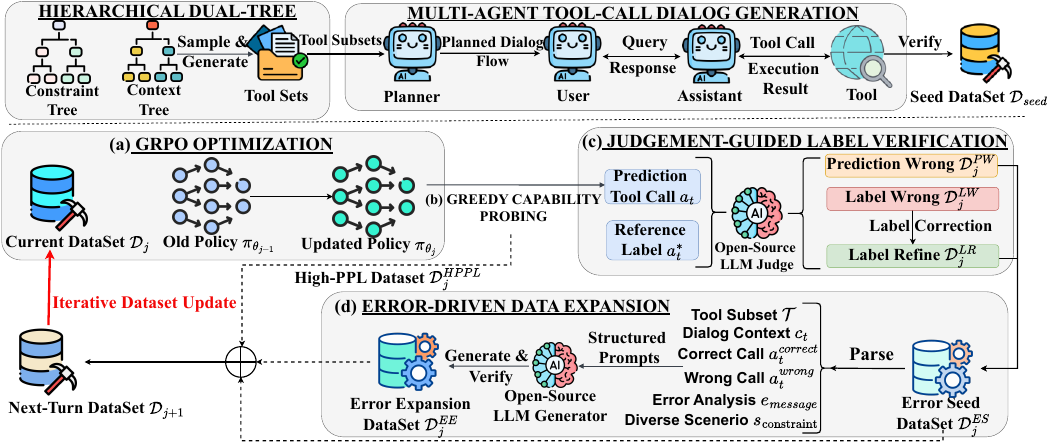}
    \vspace{0.5em}
    \caption{The overall closed-loop automatic pipeline of \textbf{\modelname}, which couples (a) GRPO optimization, (b) Greedy Capacity Probing, (c) Judgement-Guided Label Verification, and (d) Error-Driven Data Expansion for iterative tool-use enhancement.}
    \label{Fig:LoopTool}
\end{figure}



\subsection{Hierarchical Dual-Tree Guided API Synthesis}
\vspace{-0.5em}
Our tool set comprises both real-world APIs collected from public resources~\citep{ToolACE,APIGen,ToolLLM} and synthetically generated APIs produced via a \textit{Hierarchical Dual-Tree} method.
For each application domain, we define two complementary hierarchical structures: (i) \textbf{Context Tree} encodes the topical scope and functional granularity of the domain, from coarse categories at the root to fine-grained specializations at the leaves; (ii) \textbf{Constraint Tree} specifies structural and functional constraints for valid APIs, such as naming conventions, parameter types and counts, and output formats. To synthesize an API, we independently sample a leaf path from each tree and merge the results into a structured prompt for the LLM, ensuring that both functional intent and structural requirements are satisfied.
Rule-based validation is subsequently applied to ensure conformity and semantic coherence. Concrete examples of Context and Constraint Trees are provided in Appendix~\ref{appendix:dual_tree_examples}.

\subsection{Multi-Agent Tool-Use Dialog Generation}\label{Sec:Multi-Agent-Dialog}
The dialog generation stage incorporates two components: the Multi-Agent Dialogue Simulation and Correctness Verification for quality control.

\textbf{Multi-Agent Dialogue Simulation.} We populate the seed dataset by simulating tool-usage dialogues with four distinct roles: \textit{Planner Agent} designs coherent conversation flows based on a sampled subset of tools and a target number of dialog turns. This planning phase ensures realistic task decomposition and natural progression toward tool use. \textit{User Agent} interacts with the assistant according to the Planner’s high-level outline, generating new requests, clarifying requirements, or providing additional information such as missing parameters. \textit{Assistant Agent} selects appropriate APIs from the assigned subset, extracts candidate parameters based on the dialog context, executes tool calls, or synthesizes responses for the user. \textit{Tool Agent} processes the tool calls according to the given API definitions and produces simulated execution results. For certain domains, we integrate real executable backends to return authentic responses through actual code execution. The dialog proceeds turn-by-turn until the predefined conversation length is reached.

\textbf{Rule-based and LLM-based Verification.} All generated dialogues undergo a two-tier verification process. Rule-based verification checks 
API call syntax, parameter coverage, type matching, and adherence to schema definitions.
LLM-based evaluation leverages an open-source judge model (Qwen3‑32B) to holistically evaluate every tool call step for contextual appropriateness, logical consistency, and alignment with the user’s intent.
Only dialogues satisfying both stages are admitted into the initial seed dataset.

\section{Iterative Model Training and Data Augment}
To overcome the limitations of static data generation and support dynamically adaptive model training, we develop an automated iterative framework for tool-augmented LLM learning as shown in Figure~\ref{Fig:LoopTool}. \modelname integrates the GRPO Optimization, Greedy Capability Probing, Judgement-Guided Label Verification, and Error-Driven Data Expansion into a unified closed loop. This iterative cycle enables the model to assess its own capabilities continuously, target its weaknesses, and refine the quality of supervision data. 

\subsection{GRPO Training for Tool Calling}
\textbf{Data Format.} We construct an \textbf{initial seed tool-calling dialogue dataset} $\mathcal{D}_{\text{seed}}$ through the Automated Tool-Augmented Data Construction in Section~\ref{sec:bootstrapping}. Each multi-turn dialog sample is transformed into multiple GRPO training samples, which consist of the tuple: $(\mathcal{T},c_t,a_t^{*})$, where $t$ denotes the current turn in the dialogue, as a single conversation may contain multiple sequential tool calls. $\mathcal{T}$ denotes the set of available tools at the current step, $c_t$ represents the historical dialogue context, which can be either a single-turn user query or
a multi-turn conversation. $a_t^{*}$ is the tool call step from the conversation corresponding to the last user query. The model's output $O_t$ include two structured components: a reasoning trace wrapped within \texttt{<think>} \ldots \texttt{</think>} and the predicted tool invocation $a_t$ inside \texttt{<tool\_call>} \ldots \texttt{</tool\_call>}. A detailed specification of both the single-turn and multi-turn training formats is provided in Appendix~\ref{appendix:train_format}.

\textbf{Binary Reward Definition.}
To quantify the quality of model-generated tool calls, we adopt a \emph{Binary Reward} scheme, which serves as a simple yet effective rule-based reward function. For a given context $c_t$ and the model output $a_t$, the reward is defined as:
\begin{equation}
    r(\mathcal{T}, c_t, a_t^{*}, a_t) =
    \begin{cases} 
      1, &  \text{ToolMatch}(a_t, a^{*}_t) \\
      0, & \text{otherwise}
    \end{cases}
    \label{eq:reward_function}
\end{equation} 

\vspace{0.5em}
\noindent\textbf{GRPO Optimization.}
Given the tool sets $\mathcal{T}$ and historical dialogue $c_t$, the policy $\pi_\theta$ sample a group of candidate response $\{O_t^{1}, O_t^{2}, \ldots, O_t^{G}\}$ from the old policy $\pi_{\theta_{\text{old}}}$ and their corresponding rewards are $\{r_t^{1}, r_t^{2}, \ldots, r_t^{G}\}$. We optimizes the $\pi_\theta$ through maximizing the following objective:

\begin{equation}
\begin{aligned}\label{eq:GRPO}
     \mathcal{J}_{\text{GRPO}}(\theta) &= \mathbb{E}_{(\mathcal{T}, c_t) \sim \mathcal{D}, \{O_t^i\}_{i=1}^G \sim \pi_{\theta_{\text{old}}}} \, \frac{1}{G} \sum_{i=1}^{G} \left[ \min \left( \rho_t^i A_t^i, \, \text{clip}(\rho_t^i, 1 - \epsilon, 1 + \epsilon) A_t^i \right) - \beta \, \text{KL}(\pi_\theta \, \| \, \pi_\text{old}) \right], \\
      & \text{where } \rho_t^i = \frac{\pi_\theta(O_t^i \mid c_t, \mathcal{T})}{\pi_{\theta_\text{old}}(O_t^i \mid c_t, \mathcal{T})}, \quad  A_t^{i} = \frac{r_t^{i} - \text{mean}(\{r_t^{1}, r_t^{2}, \ldots, r_t^{G}\})}{\text{std}(\{r_t^{1}, r_t^{2}, \ldots, r_t^{G}\})}
\end{aligned}
\end{equation}
$\epsilon$ is the PPO clipping parameter, and $\beta$ controls the strength of the KL penalty.

\subsection{Greedy Capability Probing}
GRPO-based post-training often assigns near-zero advantage values to both trivially solvable and prohibitively hard samples, resulting in negligible parameter updates despite non-trivial computational costs~\citep{DAPO}. To mitigate this inefficiency, we introduce \textbf{Greedy Capability Probing} (GCP)—an offline diagnostic stage to identify samples of substantive learning value.

Given the training set $\mathcal{D}_j$ in the $j$‑th iteration, we perform deterministic greedy decoding with the current policy $\pi_{\theta_j}$ on every instance. For each tool-call sample $(\mathcal{T}, c_t, a_t^*)$, the model generates a prediction $a_t \in O_t$ via greedy search. If $a_t = a_t^*$, the sample is provisionally considered \emph{mastered} under the assumption that its label is correct. Otherwise, the quadruple $(\mathcal{T}, c_t, a_t^*; a_t)$ is passed to \textbf{Judgement-Guided Label Verification} (JGLV) for correctness assessment. To further quantify sample difficulty,  we compute sample-level perplexity~(PPL) as:
\begin{equation}\label{eq:PPL}
    \mathrm{PPL}_{(\mathcal{T}, c_t)}  
    = \exp\left(-\frac{1}{L} \sum_{i=1}^{L} \log p_\theta(o_i \mid \mathcal{T}, c_t, o_{1:i-1}) \right)
\end{equation}
where $L$ is the output length and $o_i$ denotes the $i$‑th token in the output sequence. High perplexity indicates low model confidence and suggests that the sample resides near the decision boundary, making it more valuable for continued training. In subsequent iterations, GCP selectively retains a subset of these high‑PPL cases $\mathcal{D}_j^{\text{HPPL}}$into the next-turn iteration.

\subsection{Judgement-Guided Label Verification}
To mitigate the impact of noisy synthetic annotations and integrate \emph{automatic label refinement} directly into the iterative loop, we introduce \textbf{Judgement-Guided Label Verification (JGLV)}—a structured evaluation stage that distinguishes genuine model failures from annotation errors.

In each iteration $j$, for every mismatched case $(\mathcal{T}, c_t, a_t^*; a_t)$ identified by \textbf{Greedy Capability Probing}, we organize the tool specifications $\mathcal{T}$, dialogue context $c_t$, reference label $a_t^*$ and model prediction $a_t$ into an open-source LLM—in our implementation, \emph{Qwen3-32B}~\citep{Qwen3_report}-which outputs a categorical decision: $y_{\mathrm{judge}} \in \{\texttt{PRED\_WRONG}, \texttt{LABEL\_WRONG}, \texttt{BOTH\_CORRECT}, \texttt{BOTH\_WRONG}\}$ and formatted error analysis $e_\text{message}$. Based on the judgment results, we define two key subsets of the evolving dataset: the Prediction Wrong set and the Label Wrong Set.
\begin{equation}
\begin{aligned}
     \mathcal{D}_j^{PW} &= \{ (\mathcal{T}, c_t, a_t^*; a_t) \;|\; y_{\mathrm{judge}} = \texttt{PRED\_WRONG} \} \\
    \mathcal{D}_j^{LW} &= \{ (\mathcal{T}, c_t, a_t^*, a_t) \;|\; y_{\mathrm{judge}} = \texttt{LABEL\_WRONG} \}
\end{aligned}
\end{equation}
$\mathcal{D}_j^{PW}$ are retained for retraining in the next iteration. We replace the $a_t^*$ in $\mathcal{D}_j^{LW}$ with $a_t$ to  transform the dataset into $\mathcal{D}_j^{LR}$(Refer to Appendix~\ref{appendix:JGLV} for judgement prompt and detailed samples). For samples classified as $\texttt{BOTH\_CORRECT}$, we retain only those with high-PPL into $\mathcal{D}_j^{\text{HPPL}}$. Samples  identified as 
$\texttt{BOTH\_WRONG}$ are directly discarded to avoid propagating noisy supervision.


Compared with approaches that rely on a large language model to directly regenerate or correct labels, JGLV reframes annotation refinement as a \textit{comparative judgment task}, where the judge model only determines which of two existing candidates better satisfies the task specification instead of producing a new output from scratch.
Moreover, by incorporating outputs from the evolving current policy into the judgment process, \textbf{JGLV leverages the model’s progressively improving tool‑calling competence to assist data refinement.} As training advances, the policy increasingly produces valid and high‑quality tool invocations, enabling the replacement of incorrect labels with superior model outputs. This synergy transforms label verification into a self‑reinforcing mechanism that continuously generates cleaner and more representative training data.

\subsection{Error-Driven Data Expansion} While GCP and JGLV effectively identify mismatched cases and correct noisy labels, reusing these instances without modification often yields marginal benefit (see Section~\ref{Sec:Ablation}), especially when failures arise from systematic weaknesses rather than incidental noise. To directly broaden the model’s coverage of challenging tool-use scenarios, we propose Error‑Driven Data Expansion (EDDE)—an augmentation strategy that transforms verified failure cases into structurally similar “hard” samples.

In iteration $j$, EDDE operates on the union of the $\mathcal{D}_j^{MR}$ and  $\mathcal{D}_j^{LR}$ identified by JGLV: $\mathcal{D}_j^{ES} = \mathcal{D}_j^{MR} \cup \mathcal{D}_j^{LR}$. For each error seed $(\mathcal{T}, c_t, a_t^*; a_t) \in \mathcal{D}_j^{ES}$, EDDE parses the following structured components: tool subset $\mathcal{T}$, dialog context $c_t$, correct call $a_t^{\text{correct}}$, wrong call $a_t^{\text{wrong}}$, and error analysis $e_{\text{message}}$. The generator is instructed to produce $k$ new tool‑calling samples that mirror the structural complexity of the error seed (e.g., similar argument, multi‑step dependencies). To avoid excessive similarity among the augmented samples derived from the same error seed, we additionally introduce scenario diversification constraints $s_{\text{constraint}}$. Specifically, each generation prompt is enriched with varied situational contexts—such as alternative user goals, different domain-specific constraints, or modified environmental conditions—while preserving the core challenge (Refer to Appendix~\ref{appendix:EDDE} for error generation prompt and new generated samples). All EDDE‑generated samples are subjected to the same two‑tier validation pipeline outlined in Section~\ref{Sec:Multi-Agent-Dialog}—including rule‑based and LLM‑based evaluation. Samples passing both filters are collected into: $\mathcal{D}_j^{EE} = \mathrm{Verify}\big(\mathrm{Generate}(\mathcal{D}_j^{\mathrm{ES}})\big)$. 

\textbf{Integration into the Iterative Loop.}  
At the end of iteration $j$, the training dataset for the next round is constructed by merging multiple sources identified during the current iteration:
\begin{equation}
\label{eq:next_iter_dataset}
\mathcal{D}_{j+1} 
= \mathcal{D}_{j}^{ES} 
\cup \mathcal{D}_{j}^{EE} 
\cup \mathcal{D}_{j}^{\text{HPPL}} 
\cup \mathcal{D}_{j}^{\text{Seed-new}}
\end{equation}
where $\mathcal{D}_{j}^{\text{Seed-new}}$ is a small untrained subset from the initial seed dataset $\mathcal{D}_{\text{seed}}$. This merged dataset $\mathcal{D}_{j+1}$ is then used in the subsequent GRPO training round, with the policy $\pi_{\theta_j}$ serving as the initialization. The full iteration pipeline is summarized in the Algorithm~\ref{Algorithm:LoopTool}.


\section{Experiments}
\vspace{-1em}

\begin{table}[ht]
\centering
\caption{Comprehensive evaluation of the BFCL-v3 (last updated on 2025-06-14). FC denotes that the model is tailored for functional calling. 
The best results in each category are highlighted in bold, while the second-best are underlined.
}
\label{Table:BFCL_overall}
\resizebox{1.0\linewidth}{!}{%
\begin{tabular}{cc|c|cc|c|cc}
\hline
\multirow{2}{*}{\textbf{Rank}} &
  \multirow{2}{*}{\textbf{Overall Acc}} &
  \multirow{2}{*}{\textbf{Model}} &
  \multicolumn{2}{c}{\textbf{Single-Turn}} &
  \textbf{Multi-Turn} &
  \multicolumn{2}{c}{\textbf{Hallucination}} \\
 &
   &
   &
  \textbf{Non-Live AST Acc} &
  \textbf{Live Acc} &
  \textbf{Overall Acc} &
  \textbf{Relevance} &
  \textbf{Irrelevance} \\ \hline\hline
1          & 78.45   & xLAM-2-70b-fc-r (FC)            & 88.44   & 72.95   & \textbf{75.00 }  & 66.67   & 78.91   \\ 
2          & 76.43   & xLAM-2-32b-fc-r (FC)            & \underline{89.27}   & 74.23   & 67.12   & \underline{88.89}   & 76.74   \\ \rowcolor{cyan!20}
\textbf{3} & \textbf{74.93} & \textbf{\modelname-8B (Ours)}                   & \textbf{89.52} & \textbf{84.72} & 50.88 & 61.11 & \underline{87.67} \\ 
4          & 73.57   & watt-tool-70B (FC)              & 84.06   & 77.74   & 58.87   & \textbf{94.44}   & 76.32   \\ 
5          & 72.04   & xLAM-2-8b-fc-r (FC)             & 84.40   & 66.90   & \underline{69.12}  & 77.78   & 64.34   \\ 
6          & 71.71   & GPT-4o-2024-11-20 (FC)          & 86.81   & 78.85   & 50.00   & 83.33   & 81.31   \\ 
7          & 70.42   & GPT-4o-2024-11-20 (Prompt)      & 87.67   & 79.88   & 43.00   & 72.22   & 85.36   \\ 
8          & 70.32   & GPT-4.5-Preview-2025-02-27 (FC) & 86.12   & 79.34   & 45.38   & 66.67   & 83.64   \\ 
9          & 69.25   & Qwen3-32B (FC)                  & 88.90   & 77.83   & 43.12   & 72.22   & 75.79   \\ 
10         & 68.89   & GPT-4.1-2025-04-14 (FC)         & 85.42   & 79.92   & 40.50   & 77.78   & 85.95   \\ 
11         & 68.73   & ToolACE-2-8B (FC)               & 87.58   & 80.05   & 37.00   & 72.22   & \textbf{90.11}   \\ 
\multicolumn{8}{c}{\dots\; (Ranks 12--18 omitted for brevity)} \\
19         & 66.34   & Qwen3-8B (FC)                   & 88.81   & 78.54   & 33.00   & 77.78   & 79.08   \\
20         & 65.19   & Qwen3-8B (FC, \textbf{self-host})        & 87.06  & 78.50   & 31.25   & 77.78   & 78.74   \\
\hline
\end{tabular}
}
\end{table}

\begin{table}[ht]
\centering
\caption{Comprehensive evaluation of ACEBench for English Data (last updated on 2025-07-21). LoopTool-8B (Ours) achieves the best result in the 8B scale.}
\label{Table:ACEBench_overall}
\resizebox{1.0\linewidth}{!}{%
\begin{tabular}{cccccccccc}
\hline
\multirow{2}{*}{\textbf{Model}} &
  \multicolumn{6}{c}{\textbf{Normal}} &
  \multirow{2}{*}{\textbf{Special}} &
  \multirow{2}{*}{\textbf{Agent}} &
  \multirow{2}{*}{\textbf{Overall}} \\ \cline{2-7}
 &
  \textbf{Atom} &
  \textbf{Single-Turn} &
  \textbf{Multi-Turn} &
  \textbf{Similar API} &
  \textbf{Perference} &
  \textbf{Summary} &
   &
   &
   \\ \hline
\multicolumn{10}{c}{\textbf{Closed-Source Large Language Models}}                                  \\ \hline
GPT-4o                     & 90.0 & 78.0 & 68.0 & 80.0 & 78.0 & 82.5 & 92.7 & 56.0 & \textbf{81.1} \\
Gemini-2.5-Pro-05-06                & 83.7 & 73.5 & 61.0 & 72.0 & 58.0 & 75.1 & 90.7 & 52.5 & \textbf{75.8} \\
Qwen-Max                   & 88.0 & 75.0 & 61.0 & 74.0 & 82.0 & 79.7 & 74.0 & 60.0 & \textbf{75.1} \\
GPT-4o-Mini                & 84.3 & 73.5 & 59.0 & 74.0 & 72.0 & 76.4 & 76.7 & 27.5 & \textbf{68.9} \\
Gemini-1.5-Pro             & 82.3 & 73.0 & 61.0 & 74.0 & 72.0 & 75.7 & 77.3 & 26.0 & \textbf{68.5} \\
Claude-3-5-Sonnet          & 66.7 & 64.0 & 46.0 & 58.0 & 68.0 & 62.2 & 72.7 & 44.0 & \textbf{62.2} \\
Doubao-Pro-32k             & 75.3 & 58.0 & 52.0 & 70.0 & 54.0 & 66.3 & 50.7 & 26.5 & \textbf{56.0} \\ \hline
\multicolumn{10}{c}{\textbf{Open-Source Large Language Models}}                                    \\ \hline
Kimi-k2-0711 & 87.0 & 78.5 & 62.0 & 70.0 & 74.0 & 78.9 & 81.3 & 65.0 & \textbf{77.4} \\
Qwen2.5-Coder-32B-Instruct & 86.0 & 73.5 & 59.0 & 76.0 & 72.0 & 77.4 & 80.0 & 50.0 & \textbf{73.9} \\
\rowcolor{cyan!20}
\textbf{\modelname-8B (Ours)} &
  \textbf{86.0} &
  \textbf{76.0} &
  \textbf{58.0} &
  \textbf{74.0} &
  \textbf{78.0} &
  \textbf{78.0} &
  \textbf{80.7} &
  \textbf{43.3} &
  \textbf{73.4} \\
ToolACE-2.5-Llama-3.1-8B &  87.7 & 75.5 & 62.0 & 74.0 & 66.0 & 78.3 & 76.0 & 35.9 & \textbf{71.1} \\
DeepSeek-V3                & 88.0 & 77.5 & 63.0 & 76.0 & 78.0 & 80.3 & 72.7 & 34.0 & \textbf{71.1} \\
Qwen2.5-72B-Instruct       & 81.3 & 74.5 & 64.0 & 76.0 & 80.0 & 76.8 & 74.0 & 37.5 & \textbf{70.0} \\
Qwen3-8B                   & 80.3 & 68.5 & 52.0 & 70.0 & 58.0 & 70.9 & 78.0 & 34.2 & \textbf{67.1} \\
Llama-3.1-70B-Instruct     & 83.7 & 71.5 & 61.0 & 74.0 & 66.0 & 75.6 & 29.3 & 41.0 & \textbf{57.9} \\
Qwen2.5-7B-Instruct        & 70.3 & 57.0 & 49.0 & 62.0 & 58.0 & 62.8 & 49.3 & 15.0 & \textbf{51.8} \\
Qwen2.5-Coder-7B-Instruct  & 73.3 & 63.5 & 52.0 & 70.0 & 58.0 & 66.6 & 25.3 & 18.5 & \textbf{48.1} \\ \hline
\end{tabular}%
}
\end{table}

\subsection{Experiment Setup}
\vspace{-1em}
\textbf{Benchmarks.} We evaluate LoopTool by training LLMs with our data generation pipeline, using the open-source Qwen3-8B~\citep{Qwen3_report} as the backbone under pure RL fine-tuning. Experiments are conducted on two representative benchmarks: BFCL-v3~\citep{BFCL} and ACEBench~\citep{ACEBench}, which provide comprehensive, executable function-call tasks for assessing function invocation capability. We also perform ablation studies to examine the contribution of individual modules. Benchmark details and evaluation metrics are provided in Appendix~\ref{appendix:benchmark}.

\textbf{Implementation Details.} GRPO training is implemented with the open-source RL library Verl~\citep{verl}, using a batch size of 128 and a learning rate of $1 \times 10^{-6}$. Each iteration trains for two epochs, resetting optimizer parameters while initializing from the previous checkpoint. To promote exploration, the actor rollout temperature is fixed at $1.0$, with both entropy coefficient and KL weight set to $0$. We apply the Clip-Higher~\citep{DAPO} strategy, increasing 
$\mathcal{E}_{high}$ from 0.2 to 0.28 to encourage generation of high-entropy, low-probability tokens. In EDDE, $k$ is set to 4. Full hyperparameters are listed in Appendix~\ref{appendix:hyperparameter}.

\vspace{-1em}

\subsection{Overall Performance Analysis}
\textbf{Result on BFCL and ACEBench.} 
We compare LoopTool-8B model with various representation models in BFCL~\citep{BFCL} and ACEBench~\citep{ACEBench}. We adopt the official evaluation script and report the average accuracy across categories. The results are summarized in Table~\ref {Table:BFCL_overall} and Table~\ref{Table:ACEBench_overall}, respectively. On both BFCL‑v3 and ACEBench leaderboards, LoopTool-8B achieves SOTA performance among all 8B‑scale open-source models and exceeds several larger counterparts. In BFCL‑v3 (Table~\ref{Table:BFCL_overall}), our model attains an overall accuracy of \textbf{74.93\%}, ranking third across all models and surpassing the original Qwen3‑8B by \textbf{+8.59 points}, with the highest Single‑Turn and Live execution accuracy. Remarkably, LoopTool‑8B also outperforms the 32B‑scale Qwen3 model—used as both the data generator and judge in our pipeline, demonstrating the capability amplification achieved through our model‑aware iterative data evolution. On ACEBench (Table \ref{Table:ACEBench_overall}), LoopTool‑8B obtains an overall accuracy of \textbf{73.4\%}, improving over Qwen3‑8B by \textbf{+6.3 points} and consistently delivering balanced gains across diverse evaluation categories.

\subsection{Iterative Details and Analysis}

\subsubsection{Iterative Dataset Distribution}

\begin{table}[htbp]
    \centering
    \renewcommand{\arraystretch}{0.85}
    \caption{Distribution of samples across iterative datasets in our LoopTool framework.  }
\label{Table:Data_Distribution}
    \resizebox{0.9\linewidth}{!}{
    \begin{tabular}{@{}cc|cccc@{}}
        \toprule
        \textbf{} & \textbf{\# Total} & \# $\mathcal{D}_j^{\text{ES}}$ &  \# $\mathcal{D}_j^{\text{EE}}$ & \# $\mathcal{D}_j^{\text{HPPL}}$   &  \# $\mathcal{D}_j^{\text{Seed-new}}$ \\ \midrule
        $\mathcal{D}_1$ & 18304 & 0 & 0 & 0 & 18304 (100\%)  \\
       $\mathcal{D}_2$ & 18304 & 1919 (10.48\%) & 6566 (35.87\%) & 4187 (22.98\%) & 5632 (30.77\%)  \\
        $\mathcal{D}_3$ & 18304 & 3386 (18.50\%) & 8066 (44.07\%) & 4036 (22.06\%) & 2816 (15.38\%) \\
        $\mathcal{D}_4$ & 18304 & 3731 (20.38\%) & 8169 (44.63\%) & 4996 (27.29\%) & 1408 (7.69\%)\\ 
        \bottomrule
    \end{tabular}
}
\end{table}
The initial seed dataset $\mathcal{D}_{\text{seed}}$ includes $28k$ tool call samples. 
The corpus $\mathcal{D}_{j+1}$ at iteration $j+1$ is constructed from four primary sources as illustrated in Eq (\ref{eq:next_iter_dataset}). $\mathcal{D}_{j}^{\text{Seed-new}}$ means the untrained new seed samples randomly drawn from the seed dataset $\mathcal{D}_{\text{seed}}$. In each iteration, we gradually reduce the proportion of untrained seed samples, ensuring that each training round incorporates newly generalized queries, while gradually converging on increasingly challenging samples.
The detailed data statistics are presented in Table~\ref{Table:Data_Distribution}.

\subsubsection{Performance Analysis of Iterative Training Framework}

\begin{figure}
    \centering
    \includegraphics[width=0.9\linewidth]{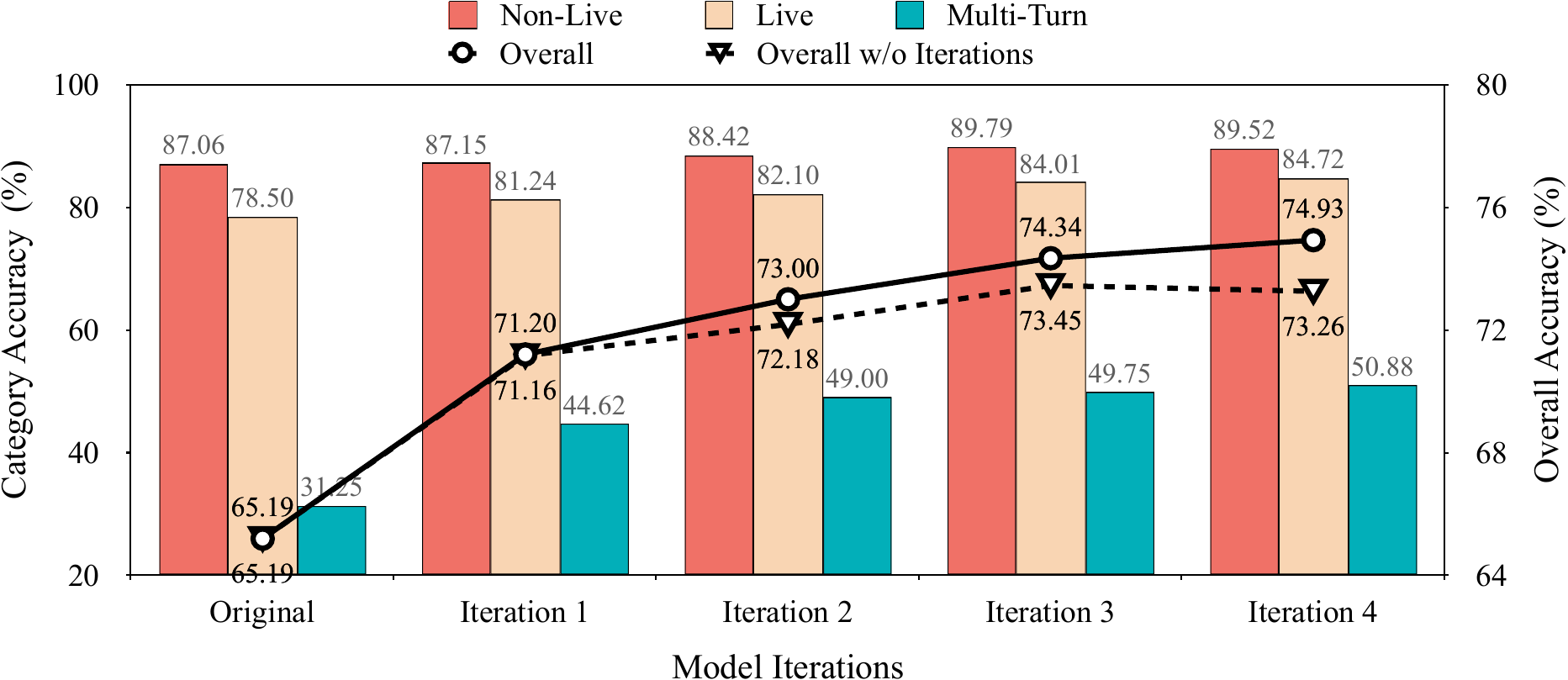}
    \caption{The Iterative Performance across four iterations evaluated in BFCL-v3. The left y-axis represents Category Acc (bar chart), while the right y-axis denotes  Overall Acc (line chart).``Overall w/o Iterations" refers to the result obtained under the same number of iteration steps, where we train solely on the initial seed dataset $\mathcal{D}_{\text{seed}}$.}
    \label{Fig:Iterative_Performance}
\end{figure}

We evaluate the effectiveness of the iterative training paradigm against conventional static data generation.
As shown in Figure~\ref{Fig:Iterative_Performance}, the proposed LoopTool framework delivers consistent gains in tool-calling accuracy across iterations. Starting from the initial model (“Original”), each iteration leverages the closed-loop data evolution to uncover and remedy model deficiencies, leading to steady improvements. In contrast, the static “Overall w/o Iterations” setting produces substantially smaller improvements. Without the injection of newly synthesized hard cases or label refinements, the model rapidly saturates on the limited supervision, exhausting the information content of $\mathcal{D}_{\text{seed}}$. Improvements plateau by Iteration~2 and decline after Iteration~3, indicating overfitting and a growing mismatch between the fixed training distribution and the model’s evolving inference behavior.

\subsection{Ablation Study} \label{Sec:Ablation}
\begin{table}[ht]
\centering
\caption{ We conduct the corresponding ablation experiments in Iteration 2 and Iteration 3, employing the data variants of $\mathcal{D}_2$ and $\mathcal{D}_3$. Overall accuracy and per-category accuracy are reported.}
\label{Table:Ablation_Study}
\begin{tabular}{lcccc}
\toprule
\textbf{Configuration} & \textbf{Overall Acc} & \textbf{Non-Live Acc} & \textbf{Live Acc} & \textbf{Multi-Turn Acc} \\
\midrule
Iteration~1 ($\mathcal{D}_1$) & 71.20 & 87.10 & 81.34 & 44.62 \\ \hline\hline
Iteration~2 ($\mathcal{D}_2$) & \textbf{73.00} & \textbf{88.42} & 82.10 & \textbf{49.00} \\
\quad w/o High-PPL & 72.31 & 88.17 & 81.59 & 46.25 \\
\quad w/o JGLV & 71.30 & 87.90 & 82.05 & 43.88 \\
\quad Remove EDDE & 71.50 & 88.06 & 81.47 & 45.00 \\
\quad HighPPL-Replace & 72.50 & 88.10 & \textbf{82.36} & 47.88 \\ 
\quad Error-Seed Repetition &  72.38 & 88.40 & 81.87 & 46.88 \\ 
\hline\hline
Iteration~3 ($\mathcal{D}_3$) & \textbf{74.34} & \textbf{89.79} & \textbf{84.01} & \textbf{49.75} \\
\quad w/o High-PPL & 73.50 & 89.12 & 82.79 & 48.90  \\
\quad w/o JGLV & 72.61 & 89.17 & 82.59 & 46.25 \\
\quad Remove EDDE & 73.12 & 88.75  & 82.45 & 48.75 \\
\quad HighPPL-Replace & 73.28 & 89.40 & 83.96 & 46.88 \\
\quad Error-Seed Repetition &  73.43 & 88.15 & 83.74 & 48.38 \\
\bottomrule
\end{tabular}
\end{table}

\begin{figure}[t]
    \centering
    \begin{minipage}[t]{0.48\linewidth}
        \centering
        \includegraphics[width=\linewidth]{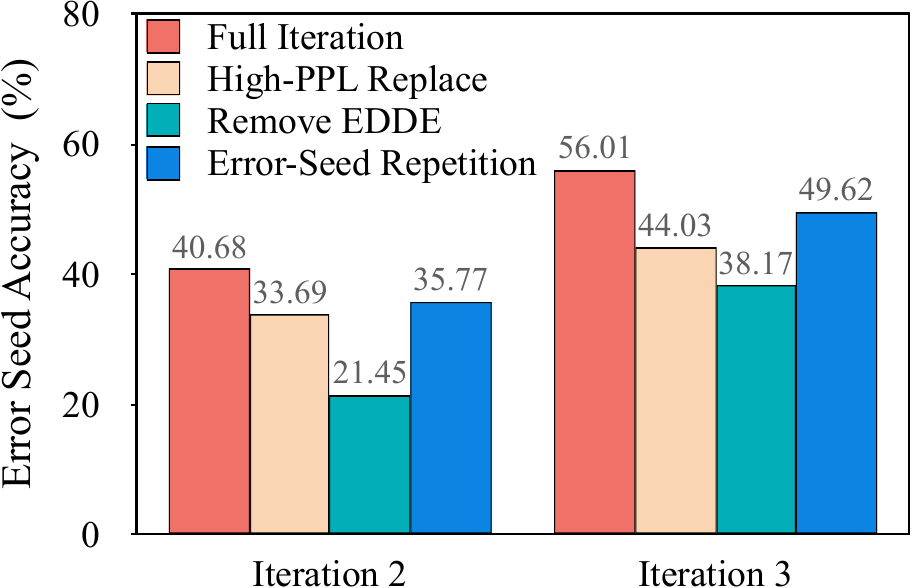}
        \caption{The Prediction Accuracy of Error Seed across iterations.}
        \label{Fig:edde_ablation}
    \end{minipage}
    \hfill
    \begin{minipage}[t]{0.48\linewidth}
        \centering
\includegraphics[width=\linewidth]{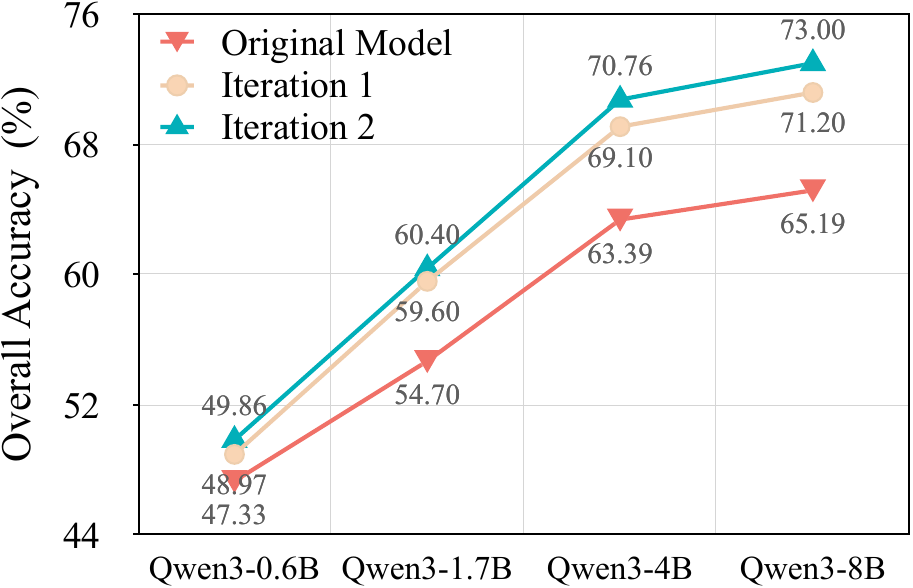}
        \caption{Scaling performance with different model sizes.}
    \label{Fig:scaling_performance}
    \end{minipage}
\end{figure}

To assess the contributions of each key component in LoopTool, we perform ablation experiments on BFCL-v3. Specifically, we design the following variants: (i) \textbf{w/o High-PPL}: Replace $\mathcal{D}_j^{\text{HPPL}}$ with randomly samples that the model predicted correctly; (ii) \textbf{w/o JGLV}: Skip verification and treat all mismatches (\(a_t \neq a_t^{*}\)) as model errors; keep original labels without refinement. (iii) \textbf{Remove EDDE}: Drop $\mathcal{D}_j^{EE}$ entirely; (iv) \textbf{HighPPL-Replace}: Replace $\mathcal{D}_j^{EE}$ with an equal number of high‑PPL samples selected via GCP; (v) \textbf{Error-Seed Repetition}: Remove $\mathcal{D}_j^{EE}$ and duplicate $\mathcal{D}_j^{ES}$ to match data scale. From the results in Table~\ref{Table:Ablation_Study}, several key observations can be made:
From the results in Table~\ref{Table:Ablation_Study}, several key observations can be made: 
\vspace{-8pt}
\begin{itemize}[leftmargin=10pt]
    \item \textbf{Importance of high-PPL samples}. \textbf{w/o High-PPL} lead to consistent accuracy drops, especially in Multi-Turn cases. Even replacing EDDE samples with high-PPL ones (\textbf{HighPPL-Replace}) sustains performance close to full configurations, confirming that high-PPL cases—though previously predicted correctly—lie near decision boundaries of current policy and drive further refinement, in line with recent works~\citep{SvS,rstar2agent}.
    \vspace{-0.5em}
    \item \textbf{Necessity of JGLV.} 
Skipping verification (\textbf{w/o JGLV}) significantly degrades accuracy, confirming that noisy or erroneous labels can mislead training. Without label refinement, such errors persist and even propagate when used by EDDE to generate variants, exacerbating noise in subsequent iterations. 
\vspace{-0.5em}
\item \textbf{Effectiveness of EDDE} The three variants of \textbf{w/o EDDE} in both Iteration~2 and Iteration~3, result in consistent drops in overall accuracy. To further quantify the direct contribution of EDDE-originated samples, we compare the three variants with full configuration, testing the accuracy exclusively on this subset of historically wrong cases, with results shown in Figure \ref{Fig:edde_ablation}. The result illustrates that simply re-training the model on the original erroneous seeds is insufficient for the model to master these difficult cases effectively. In contrast, EDDE synthesizes structurally similar, error-informed variants that preserve the underlying challenges of the original failure cases while offering additional diversity. This targeted augmentation enables the model to acquire the relevant patterns more reliably, thereby improving its performance on the original hard seeds.
\end{itemize}

\subsection{Scaling Performance with Model Size}  
We evaluate LoopTool across backbone models from 0.6B to 8B parameters, measuring BFCL-v3 accuracy over two training iterations (Figure~\ref{Fig:scaling_performance}). Larger models consistently achieve higher accuracy in both the initial (\textit{Iteration 1}) and refined (\textit{Iteration 2}) stages, with greater absolute improvements in the second iteration. Specifically, the 0.6B model gains only $+0.70$ points, whereas the 8B model achieves $+1.80$ points. This scaling trend stems from GRPO-based post-training, which depends on the model’s ability to discover correct tool-use trajectories during rollout exploration. Larger models tend to identify such trajectories earlier, thereby amplifying the benefits of iterative refinement.

\subsection{Generalization Ability Evaluation}
\begin{table}[ht]
\centering
\setlength{\tabcolsep}{4pt}
\caption{Generalization benchmark performance comparison between vanilla Qwen3‑8B and LoopTool‑8B. Bold indicates the better score for each task.}
\label{tab:generalization}
\begin{tabular}{lcccccc}
\toprule
\textbf{Model} & \textbf{MMLU‑redux} & \textbf{IFEval} & \textbf{LiveCodeBench} & \textbf{Math‑500} & \textbf{AIME24} & \textbf{AIME25} \\
\midrule
\textbf{Qwen3‑8B}    & \textbf{87.72} & 83.30 & 42.31 & 91.40 & 60.00   & 56.67 \\
\textbf{LoopTool‑8B} & 87.37 & \textbf{84.70} & \textbf{46.15} & \textbf{92.60} & \textbf{70.00} & \textbf{66.67} \\
\bottomrule
\end{tabular}
\end{table}

Beyond tool-use performance, we evaluate whether the LoopTool‑8B model maintains or improves generalization to non-tool-related domains. We compare LoopTool‑8B with the vanilla Qwen3‑8B~\citep{Qwen3_report} across six representative benchmarks: MMLU-redux~\citep{MMLU_Redux}, IFEVAL~\citep{IFEval}, LiveCodeBench~\citep{LiveCodeBench}, Math-500~\citep{Math-500}, AIME24 and AIME25~\citealp{AIMEProblems}.

LoopTool‑8B consistently matches or surpasses Qwen3‑8B across all domains, with notable improvements in instruction-following (+1.40 on \texttt{IFEval}), code generation (+3.84 on \texttt{LiveCodeBench}), and mathematics (+1.20 on \texttt{Math‑500}, and gains on both \texttt{AIME} sets. These results indicate that the proposed iterative, model-aware data refinement and training paradigm avoids overfitting to tool-calling tasks. Instead, it fosters improved general reasoning and problem‑solving skills, enhancing the model’s capacity to generalize across diverse scenarios.

\section{Conclusion and Limitation}
We present \textbf{LoopTool}, a fully automated, model‑aware pipeline that integrates data synthesis, label refinement, and GRPO‑based post‑training into a closed loop to enhance tool‑augmented LLMs. This unified process yields progressively cleaner and more challenging data without dependence on costly closed‑source APIs, leveraging a single open‑source model for both judgment and generation. Experiments show that our 8B‑scale model trained with LoopTool surpasses its own 32B‑scale generator, highlighting the amplifying effect of iterative, model‑aware data evolution. Nonetheless, LoopTool currently operates as an offline iterative framework, in which training data evolution cannot occur concurrently with the model’s training process. LoopTool is also strictly serial per iteration, with subsequent rounds only beginning after the previous iteration finishes. Future work will explore online or streaming variants, as well as parallelized iteration schemes, to enable faster and more adaptive data–model co‑evolution.

\bibliography{iclr2026_conference}

@article{Tool_Learning_Survey,
   title={Tool learning with large language models: a survey},
   volume={19},
   ISSN={2095-2236},
   url={http://dx.doi.org/10.1007/s11704-024-40678-2},
   DOI={10.1007/s11704-024-40678-2},
   number={8},
   journal={Frontiers of Computer Science},
   publisher={Springer Science and Business Media LLC},
   author={Qu, Changle and Dai, Sunhao and Wei, Xiaochi and Cai, Hengyi and Wang, Shuaiqiang and Yin, Dawei and Xu, Jun and Wen, Ji-rong},
   year={2025},
   month=jan }

@misc{ToolLLM,
      title={ToolLLM: Facilitating Large Language Models to Master 16000+ Real-world APIs}, 
      author={Yujia Qin and Shihao Liang and Yining Ye and Kunlun Zhu and Lan Yan and Yaxi Lu and Yankai Lin and Xin Cong and Xiangru Tang and Bill Qian and Sihan Zhao and Lauren Hong and Runchu Tian and Ruobing Xie and Jie Zhou and Mark Gerstein and Dahai Li and Zhiyuan Liu and Maosong Sun},
      year={2023},
      eprint={2307.16789},
      archivePrefix={arXiv},
      primaryClass={cs.AI},
      url={https://arxiv.org/abs/2307.16789}, 
}

@misc{tau_bench,
      title={$\tau$-bench: A Benchmark for Tool-Agent-User Interaction in Real-World Domains}, 
      author={Shunyu Yao and Noah Shinn and Pedram Razavi and Karthik Narasimhan},
      year={2024},
      eprint={2406.12045},
      archivePrefix={arXiv},
      primaryClass={cs.AI},
      url={https://arxiv.org/abs/2406.12045}, 
}

@misc{ToolRL,
      title={ToolRL: Reward is All Tool Learning Needs}, 
      author={Cheng Qian and Emre Can Acikgoz and Qi He and Hongru Wang and Xiusi Chen and Dilek Hakkani-Tür and Gokhan Tur and Heng Ji},
      year={2025},
      eprint={2504.13958},
      archivePrefix={arXiv},
      primaryClass={cs.LG},
      url={https://arxiv.org/abs/2504.13958}, 
}

@misc{ToolFormer,
      title={Toolformer: Language Models Can Teach Themselves to Use Tools}, 
      author={Timo Schick and Jane Dwivedi-Yu and Roberto Dessì and Roberta Raileanu and Maria Lomeli and Luke Zettlemoyer and Nicola Cancedda and Thomas Scialom},
      year={2023},
      eprint={2302.04761},
      archivePrefix={arXiv},
      primaryClass={cs.CL},
      url={https://arxiv.org/abs/2302.04761}, 
}

@misc{ToolACE,
      title={ToolACE: Winning the Points of LLM Function Calling}, 
      author={Weiwen Liu and Xu Huang and Xingshan Zeng and Xinlong Hao and Shuai Yu and Dexun Li and Shuai Wang and Weinan Gan and Zhengying Liu and Yuanqing Yu and Zezhong Wang and Yuxian Wang and Wu Ning and Yutai Hou and Bin Wang and Chuhan Wu and Xinzhi Wang and Yong Liu and Yasheng Wang and Duyu Tang and Dandan Tu and Lifeng Shang and Xin Jiang and Ruiming Tang and Defu Lian and Qun Liu and Enhong Chen},
      year={2025},
      eprint={2409.00920},
      archivePrefix={arXiv},
      primaryClass={cs.LG},
      url={https://arxiv.org/abs/2409.00920}, 
}

@misc{TravelPlanner,
      title={TravelPlanner: A Benchmark for Real-World Planning with Language Agents}, 
      author={Jian Xie and Kai Zhang and Jiangjie Chen and Tinghui Zhu and Renze Lou and Yuandong Tian and Yanghua Xiao and Yu Su},
      year={2024},
      eprint={2402.01622},
      archivePrefix={arXiv},
      primaryClass={cs.CL},
      url={https://arxiv.org/abs/2402.01622}, 
}

@misc{ReSearch,
      title={ReSearch: Learning to Reason with Search for LLMs via Reinforcement Learning}, 
      author={Mingyang Chen and Tianpeng Li and Haoze Sun and Yijie Zhou and Chenzheng Zhu and Haofen Wang and Jeff Z. Pan and Wen Zhang and Huajun Chen and Fan Yang and Zenan Zhou and Weipeng Chen},
      year={2025},
      eprint={2503.19470},
      archivePrefix={arXiv},
      primaryClass={cs.AI},
      url={https://arxiv.org/abs/2503.19470}, 
}

@misc{ToolN1,
      title={Nemotron-Research-Tool-N1: Exploring Tool-Using Language Models with Reinforced Reasoning}, 
      author={Shaokun Zhang and Yi Dong and Jieyu Zhang and Jan Kautz and Bryan Catanzaro and Andrew Tao and Qingyun Wu and Zhiding Yu and Guilin Liu},
      year={2025},
      eprint={2505.00024},
      archivePrefix={arXiv},
      primaryClass={cs.CL},
      url={https://arxiv.org/abs/2505.00024}, 
}

@misc{chen2025Button,
      title={Facilitating Multi-turn Function Calling for LLMs via Compositional Instruction Tuning}, 
      author={Mingyang Chen and Haoze Sun and Tianpeng Li and Fan Yang and Hao Liang and Keer Lu and Bin Cui and Wentao Zhang and Zenan Zhou and Weipeng Chen},
      year={2025},
      eprint={2410.12952},
      archivePrefix={arXiv},
      primaryClass={cs.CL},
      url={https://arxiv.org/abs/2410.12952}, 
}

@misc{APIGen,
      title={APIGen: Automated Pipeline for Generating Verifiable and Diverse Function-Calling Datasets}, 
      author={Zuxin Liu and Thai Hoang and Jianguo Zhang and Ming Zhu and Tian Lan and Shirley Kokane and Juntao Tan and Weiran Yao and Zhiwei Liu and Yihao Feng and Rithesh Murthy and Liangwei Yang and Silvio Savarese and Juan Carlos Niebles and Huan Wang and Shelby Heinecke and Caiming Xiong},
      year={2024},
      eprint={2406.18518},
      archivePrefix={arXiv},
      primaryClass={cs.CL},
      url={https://arxiv.org/abs/2406.18518}, 
}

@misc{APIGen-MT,
      title={APIGen-MT: Agentic Pipeline for Multi-Turn Data Generation via Simulated Agent-Human Interplay}, 
      author={Akshara Prabhakar and Zuxin Liu and Ming Zhu and Jianguo Zhang and Tulika Awalgaonkar and Shiyu Wang and Zhiwei Liu and Haolin Chen and Thai Hoang and Juan Carlos Niebles and Shelby Heinecke and Weiran Yao and Huan Wang and Silvio Savarese and Caiming Xiong},
      year={2025},
      eprint={2504.03601},
      archivePrefix={arXiv},
      primaryClass={cs.CL},
      url={https://arxiv.org/abs/2504.03601}, 
}

@misc{tau2bench,
      title={$\tau^2$-Bench: Evaluating Conversational Agents in a Dual-Control Environment}, 
      author={Victor Barres and Honghua Dong and Soham Ray and Xujie Si and Karthik Narasimhan},
      year={2025},
      eprint={2506.07982},
      archivePrefix={arXiv},
      primaryClass={cs.AI},
      url={https://arxiv.org/abs/2506.07982}, 
}

@misc{survey_elec_automation,
      title={A Survey of Research in Large Language Models for Electronic Design Automation}, 
      author={Jingyu Pan and Guanglei Zhou and Chen-Chia Chang and Isaac Jacobson and Jiang Hu and Yiran Chen},
      year={2025},
      eprint={2501.09655},
      archivePrefix={arXiv},
      primaryClass={cs.LG},
      url={https://arxiv.org/abs/2501.09655}, 
}

@misc{ToolAlpaca,
      title={ToolAlpaca: Generalized Tool Learning for Language Models with 3000 Simulated Cases}, 
      author={Qiaoyu Tang and Ziliang Deng and Hongyu Lin and Xianpei Han and Qiao Liang and Boxi Cao and Le Sun},
      year={2023},
      eprint={2306.05301},
      archivePrefix={arXiv},
      primaryClass={cs.CL},
      url={https://arxiv.org/abs/2306.05301}, 
}

@misc{RL_enhanced_LLM_survey,
      title={Reinforcement Learning Enhanced LLMs: A Survey}, 
      author={Shuhe Wang and Shengyu Zhang and Jie Zhang and Runyi Hu and Xiaoya Li and Tianwei Zhang and Jiwei Li and Fei Wu and Guoyin Wang and Eduard Hovy},
      year={2025},
      eprint={2412.10400},
      archivePrefix={arXiv},
      primaryClass={cs.CL},
      url={https://arxiv.org/abs/2412.10400}, 
}

@misc{DeepSeekMath,
      title={DeepSeekMath: Pushing the Limits of Mathematical Reasoning in Open Language Models}, 
      author={Zhihong Shao and Peiyi Wang and Qihao Zhu and Runxin Xu and Junxiao Song and Xiao Bi and Haowei Zhang and Mingchuan Zhang and Y. K. Li and Y. Wu and Daya Guo},
      year={2024},
      eprint={2402.03300},
      archivePrefix={arXiv},
      primaryClass={cs.CL},
      url={https://arxiv.org/abs/2402.03300}, 
}

@misc{ACEBench,
      title={ACEBench: Who Wins the Match Point in Tool Usage?}, 
      author={Chen Chen and Xinlong Hao and Weiwen Liu and Xu Huang and Xingshan Zeng and Shuai Yu and Dexun Li and Shuai Wang and Weinan Gan and Yuefeng Huang and Wulong Liu and Xinzhi Wang and Defu Lian and Baoqun Yin and Yasheng Wang and Wu Liu},
      year={2025},
      eprint={2501.12851},
      archivePrefix={arXiv},
      primaryClass={cs.CL},
      url={https://arxiv.org/abs/2501.12851}, 
}

@inproceedings{
BFCL,
title={The Berkeley Function Calling Leaderboard ({BFCL}): From Tool Use to Agentic Evaluation of Large Language Models},
author={Shishir G Patil and Huanzhi Mao and Fanjia Yan and Charlie Cheng-Jie Ji and Vishnu Suresh and Ion Stoica and Joseph E. Gonzalez},
booktitle={Forty-second International Conference on Machine Learning},
year={2025},
url={https://openreview.net/forum?id=2GmDdhBdDk}
}

@misc{RLHF,
      title={Training language models to follow instructions with human feedback}, 
      author={Long Ouyang and Jeff Wu and Xu Jiang and Diogo Almeida and Carroll L. Wainwright and Pamela Mishkin and Chong Zhang and Sandhini Agarwal and Katarina Slama and Alex Ray and John Schulman and Jacob Hilton and Fraser Kelton and Luke Miller and Maddie Simens and Amanda Askell and Peter Welinder and Paul Christiano and Jan Leike and Ryan Lowe},
      year={2022},
      eprint={2203.02155},
      archivePrefix={arXiv},
      primaryClass={cs.CL},
      url={https://arxiv.org/abs/2203.02155}, 
}

@misc{DPO,
      title={Direct Preference Optimization: Your Language Model is Secretly a Reward Model}, 
      author={Rafael Rafailov and Archit Sharma and Eric Mitchell and Stefano Ermon and Christopher D. Manning and Chelsea Finn},
      year={2024},
      eprint={2305.18290},
      archivePrefix={arXiv},
      primaryClass={cs.LG},
      url={https://arxiv.org/abs/2305.18290}, 
}

@misc{SimPO,
      title={SimPO: Simple Preference Optimization with a Reference-Free Reward}, 
      author={Yu Meng and Mengzhou Xia and Danqi Chen},
      year={2024},
      eprint={2405.14734},
      archivePrefix={arXiv},
      primaryClass={cs.CL},
      url={https://arxiv.org/abs/2405.14734}, 
}

@misc{lazaridou2022internetaugmentedlanguagemodelsfewshot,
      title={Internet-augmented language models through few-shot prompting for open-domain question answering}, 
      author={Angeliki Lazaridou and Elena Gribovskaya and Wojciech Stokowiec and Nikolai Grigorev},
      year={2022},
      eprint={2203.05115},
      archivePrefix={arXiv},
      primaryClass={cs.CL},
      url={https://arxiv.org/abs/2203.05115}, 
}

@misc{wang2024executablecodeactionselicit,
      title={Executable Code Actions Elicit Better LLM Agents}, 
      author={Xingyao Wang and Yangyi Chen and Lifan Yuan and Yizhe Zhang and Yunzhu Li and Hao Peng and Heng Ji},
      year={2024},
      eprint={2402.01030},
      archivePrefix={arXiv},
      primaryClass={cs.CL},
      url={https://arxiv.org/abs/2402.01030}, 
}

@misc{hu2024visualsketchpadsketchingvisual,
      title={Visual Sketchpad: Sketching as a Visual Chain of Thought for Multimodal Language Models}, 
      author={Yushi Hu and Weijia Shi and Xingyu Fu and Dan Roth and Mari Ostendorf and Luke Zettlemoyer and Noah A Smith and Ranjay Krishna},
      year={2024},
      eprint={2406.09403},
      archivePrefix={arXiv},
      primaryClass={cs.CV},
      url={https://arxiv.org/abs/2406.09403}, 
}

@misc{ma2024mmsbenchmarkevaluatetooluse,
      title={m\&m's: A Benchmark to Evaluate Tool-Use for multi-step multi-modal Tasks}, 
      author={Zixian Ma and Weikai Huang and Jieyu Zhang and Tanmay Gupta and Ranjay Krishna},
      year={2024},
      eprint={2403.11085},
      archivePrefix={arXiv},
      primaryClass={cs.CV},
      url={https://arxiv.org/abs/2403.11085}, 
}

@misc{HuggingGPT,
      title={HuggingGPT: Solving AI Tasks with ChatGPT and its Friends in Hugging Face}, 
      author={Yongliang Shen and Kaitao Song and Xu Tan and Dongsheng Li and Weiming Lu and Yueting Zhuang},
      year={2023},
      eprint={2303.17580},
      archivePrefix={arXiv},
      primaryClass={cs.CL},
      url={https://arxiv.org/abs/2303.17580}, 
}

@article{AgentInstruct,
  title={Nonlinear discrete-time observers with physics-informed neural networks},
  author={Alvarez, Hector Vargas and Fabiani, Gianluca and Kazantzis, Nikolaos and Kevrekidis, Ioannis G and Siettos, Constantinos},
  journal={Chaos, Solitons \& Fractals},
  volume={186},
  pages={115215},
  year={2024},
  publisher={Elsevier}
}

@article{MATRIX,
  title={Synthesizing post-training data for llms through multi-agent simulation},
  author={Tang, Shuo and Pang, Xianghe and Liu, Zexi and Tang, Bohan and Ye, Rui and Jin, Tian and Dong, Xiaowen and Wang, Yanfeng and Chen, Siheng},
  journal={arXiv preprint arXiv:2410.14251},
  year={2024}
}

@article{arcadinho2024automated,
  title={Automated test generation to evaluate tool-augmented LLMs as conversational AI agents},
  author={Arcadinho, Samuel and Apar{\'\i}cio, David and Almeida, Mariana},
  journal={arXiv preprint arXiv:2409.15934},
  year={2024}
}

@article{yin2025magnet,
  title={Magnet: Multi-turn tool-use data synthesis and distillation via graph translation},
  author={Yin, Fan and Wang, Zifeng and Hsu, I and Yan, Jun and Jiang, Ke and Chen, Yanfei and Gu, Jindong and Le, Long T and Chang, Kai-Wei and Lee, Chen-Yu and others},
  journal={arXiv preprint arXiv:2503.07826},
  year={2025}
}

@misc{Qwen3_report,
      title={Qwen3 Technical Report}, 
      author={An Yang and Anfeng Li and Baosong Yang and Beichen Zhang and Binyuan Hui and Bo Zheng and Bowen Yu and Chang Gao and Chengen Huang and Chenxu Lv and Chujie Zheng and Dayiheng Liu and Fan Zhou and Fei Huang and Feng Hu and Hao Ge and Haoran Wei and Huan Lin and Jialong Tang and Jian Yang and Jianhong Tu and Jianwei Zhang and Jianxin Yang and Jiaxi Yang and Jing Zhou and Jingren Zhou and Junyang Lin and Kai Dang and Keqin Bao and Kexin Yang and Le Yu and Lianghao Deng and Mei Li and Mingfeng Xue and Mingze Li and Pei Zhang and Peng Wang and Qin Zhu and Rui Men and Ruize Gao and Shixuan Liu and Shuang Luo and Tianhao Li and Tianyi Tang and Wenbiao Yin and Xingzhang Ren and Xinyu Wang and Xinyu Zhang and Xuancheng Ren and Yang Fan and Yang Su and Yichang Zhang and Yinger Zhang and Yu Wan and Yuqiong Liu and Zekun Wang and Zeyu Cui and Zhenru Zhang and Zhipeng Zhou and Zihan Qiu},
      year={2025},
      eprint={2505.09388},
      archivePrefix={arXiv},
      primaryClass={cs.CL},
      url={https://arxiv.org/abs/2505.09388}, 
}

@article{GPT-4,
  title={Gpt-4 technical report},
  author={OpenAI},
  journal={arXiv preprint arXiv:2303.08774},
  year={2023}
}

@misc{DeepSeekR1,
      title={DeepSeek-R1: Incentivizing Reasoning Capability in LLMs via Reinforcement Learning}, 
      author={DeepSeek-AI and Daya Guo and Dejian Yang and Haowei Zhang and Junxiao Song and Ruoyu Zhang and Runxin Xu and Qihao Zhu and Shirong Ma and Peiyi Wang and Xiao Bi and Xiaokang Zhang and Xingkai Yu and Yu Wu and Z. F. Wu and Zhibin Gou and Zhihong Shao and Zhuoshu Li and Ziyi Gao and Aixin Liu and Bing Xue and Bingxuan Wang and Bochao Wu and Bei Feng and Chengda Lu and Chenggang Zhao and Chengqi Deng and Chenyu Zhang and Chong Ruan and Damai Dai and Deli Chen and Dongjie Ji and Erhang Li and Fangyun Lin and Fucong Dai and Fuli Luo and Guangbo Hao and Guanting Chen and Guowei Li and H. Zhang and Han Bao and Hanwei Xu and Haocheng Wang and Honghui Ding and Huajian Xin and Huazuo Gao and Hui Qu and Hui Li and Jianzhong Guo and Jiashi Li and Jiawei Wang and Jingchang Chen and Jingyang Yuan and Junjie Qiu and Junlong Li and J. L. Cai and Jiaqi Ni and Jian Liang and Jin Chen and Kai Dong and Kai Hu and Kaige Gao and Kang Guan and Kexin Huang and Kuai Yu and Lean Wang and Lecong Zhang and Liang Zhao and Litong Wang and Liyue Zhang and Lei Xu and Leyi Xia and Mingchuan Zhang and Minghua Zhang and Minghui Tang and Meng Li and Miaojun Wang and Mingming Li and Ning Tian and Panpan Huang and Peng Zhang and Qiancheng Wang and Qinyu Chen and Qiushi Du and Ruiqi Ge and Ruisong Zhang and Ruizhe Pan and Runji Wang and R. J. Chen and R. L. Jin and Ruyi Chen and Shanghao Lu and Shangyan Zhou and Shanhuang Chen and Shengfeng Ye and Shiyu Wang and Shuiping Yu and Shunfeng Zhou and Shuting Pan and S. S. Li and Shuang Zhou and Shaoqing Wu and Shengfeng Ye and Tao Yun and Tian Pei and Tianyu Sun and T. Wang and Wangding Zeng and Wanjia Zhao and Wen Liu and Wenfeng Liang and Wenjun Gao and Wenqin Yu and Wentao Zhang and W. L. Xiao and Wei An and Xiaodong Liu and Xiaohan Wang and Xiaokang Chen and Xiaotao Nie and Xin Cheng and Xin Liu and Xin Xie and Xingchao Liu and Xinyu Yang and Xinyuan Li and Xuecheng Su and Xuheng Lin and X. Q. Li and Xiangyue Jin and Xiaojin Shen and Xiaosha Chen and Xiaowen Sun and Xiaoxiang Wang and Xinnan Song and Xinyi Zhou and Xianzu Wang and Xinxia Shan and Y. K. Li and Y. Q. Wang and Y. X. Wei and Yang Zhang and Yanhong Xu and Yao Li and Yao Zhao and Yaofeng Sun and Yaohui Wang and Yi Yu and Yichao Zhang and Yifan Shi and Yiliang Xiong and Ying He and Yishi Piao and Yisong Wang and Yixuan Tan and Yiyang Ma and Yiyuan Liu and Yongqiang Guo and Yuan Ou and Yuduan Wang and Yue Gong and Yuheng Zou and Yujia He and Yunfan Xiong and Yuxiang Luo and Yuxiang You and Yuxuan Liu and Yuyang Zhou and Y. X. Zhu and Yanhong Xu and Yanping Huang and Yaohui Li and Yi Zheng and Yuchen Zhu and Yunxian Ma and Ying Tang and Yukun Zha and Yuting Yan and Z. Z. Ren and Zehui Ren and Zhangli Sha and Zhe Fu and Zhean Xu and Zhenda Xie and Zhengyan Zhang and Zhewen Hao and Zhicheng Ma and Zhigang Yan and Zhiyu Wu and Zihui Gu and Zijia Zhu and Zijun Liu and Zilin Li and Ziwei Xie and Ziyang Song and Zizheng Pan and Zhen Huang and Zhipeng Xu and Zhongyu Zhang and Zhen Zhang},
      year={2025},
      eprint={2501.12948},
      archivePrefix={arXiv},
      primaryClass={cs.CL},
      url={https://arxiv.org/abs/2501.12948} 
}

@misc{DAPO,
      title={DAPO: An Open-Source LLM Reinforcement Learning System at Scale}, 
      author={Qiying Yu and Zheng Zhang and Ruofei Zhu and Yufeng Yuan and Xiaochen Zuo and Yu Yue and Weinan Dai and Tiantian Fan and Gaohong Liu and Lingjun Liu and Xin Liu and Haibin Lin and Zhiqi Lin and Bole Ma and Guangming Sheng and Yuxuan Tong and Chi Zhang and Mofan Zhang and Wang Zhang and Hang Zhu and Jinhua Zhu and Jiaze Chen and Jiangjie Chen and Chengyi Wang and Hongli Yu and Yuxuan Song and Xiangpeng Wei and Hao Zhou and Jingjing Liu and Wei-Ying Ma and Ya-Qin Zhang and Lin Yan and Mu Qiao and Yonghui Wu and Mingxuan Wang},
      year={2025},
      eprint={2503.14476},
      archivePrefix={arXiv},
      primaryClass={cs.LG},
      url={https://arxiv.org/abs/2503.14476}, 
}

@inproceedings{verl, series={EuroSys ’25},
   title={HybridFlow: A Flexible and Efficient RLHF Framework},
   url={http://dx.doi.org/10.1145/3689031.3696075},
   DOI={10.1145/3689031.3696075},
   booktitle={Proceedings of the Twentieth European Conference on Computer Systems},
   publisher={ACM},
   author={Sheng, Guangming and Zhang, Chi and Ye, Zilingfeng and Wu, Xibin and Zhang, Wang and Zhang, Ru and Peng, Yanghua and Lin, Haibin and Wu, Chuan},
   year={2025},
   month=mar, pages={1279–1297},
   collection={EuroSys ’25} }

@misc{SvS,
      title={Beyond Pass@1: Self-Play with Variational Problem Synthesis Sustains RLVR}, 
      author={Xiao Liang and Zhongzhi Li and Yeyun Gong and Yelong Shen and Ying Nian Wu and Zhijiang Guo and Weizhu Chen},
      year={2025},
      eprint={2508.14029},
      archivePrefix={arXiv},
      primaryClass={cs.CL},
      url={https://arxiv.org/abs/2508.14029}, 
}

@misc{rstar2agent,
      title={rStar2-Agent: Agentic Reasoning Technical Report}, 
      author={Ning Shang and Yifei Liu and Yi Zhu and Li Lyna Zhang and Weijiang Xu and Xinyu Guan and Buze Zhang and Bingcheng Dong and Xudong Zhou and Bowen Zhang and Ying Xin and Ziming Miao and Scarlett Li and Fan Yang and Mao Yang},
      year={2025},
      eprint={2508.20722},
      archivePrefix={arXiv},
      primaryClass={cs.CL},
      url={https://arxiv.org/abs/2508.20722}, 
}

@inproceedings{MMLU_Redux,
    title = "Are We Done with {MMLU}?",
    author = "Gema, Aryo Pradipta  and
      Leang, Joshua Ong Jun  and
      Hong, Giwon  and
      Devoto, Alessio  and
      Mancino, Alberto Carlo Maria  and
      Saxena, Rohit  and
      He, Xuanli  and
      Zhao, Yu  and
      Du, Xiaotang  and
      Ghasemi Madani, Mohammad Reza  and
      Barale, Claire  and
      McHardy, Robert  and
      Harris, Joshua  and
      Kaddour, Jean  and
      Van Krieken, Emile  and
      Minervini, Pasquale",
    editor = "Chiruzzo, Luis  and
      Ritter, Alan  and
      Wang, Lu",
    booktitle = "Proceedings of the 2025 Conference of the Nations of the Americas Chapter of the Association for Computational Linguistics: Human Language Technologies (Volume 1: Long Papers)",
    month = apr,
    year = "2025",
    address = "Albuquerque, New Mexico",
    publisher = "Association for Computational Linguistics",
    url = "https://aclanthology.org/2025.naacl-long.262/",
    doi = "10.18653/v1/2025.naacl-long.262",
    pages = "5069--5096",
    ISBN = "979-8-89176-189-6"
}

@misc{IFEval,
      title={Instruction-Following Evaluation for Large Language Models}, 
      author={Jeffrey Zhou and Tianjian Lu and Swaroop Mishra and Siddhartha Brahma and Sujoy Basu and Yi Luan and Denny Zhou and Le Hou},
      year={2023},
      eprint={2311.07911},
      archivePrefix={arXiv},
      primaryClass={cs.CL},
      url={https://arxiv.org/abs/2311.07911}, 
}

@misc{LiveCodeBench,
      title={LiveCodeBench: Holistic and Contamination Free Evaluation of Large Language Models for Code}, 
      author={Naman Jain and King Han and Alex Gu and Wen-Ding Li and Fanjia Yan and Tianjun Zhang and Sida Wang and Armando Solar-Lezama and Koushik Sen and Ion Stoica},
      year={2024},
      eprint={2403.07974},
      archivePrefix={arXiv},
      primaryClass={cs.SE},
      url={https://arxiv.org/abs/2403.07974}, 
}

@misc{Math-500,
      title={Let's Verify Step by Step}, 
      author={Hunter Lightman and Vineet Kosaraju and Yura Burda and Harri Edwards and Bowen Baker and Teddy Lee and Jan Leike and John Schulman and Ilya Sutskever and Karl Cobbe},
      year={2023},
      eprint={2305.20050},
      archivePrefix={arXiv},
      primaryClass={cs.LG},
      url={https://arxiv.org/abs/2305.20050}, 
}

@online{AIMEProblems,
  title        = {{AIME}: {AIME} problems and solutions, 2025},
  url          = {https://artofproblemsolving.com/wiki/index.php/AIME_Problems_and_Solutions},
  urldate      = {2025}
}
\bibliographystyle{iclr2026_conference}

\appendix
\section{The use of Large Language Models (LLMs)}
In the research process, we employed the open-source Language model as both the Judge Model and the data Generator within our proposed LoopTool framework. During manuscript preparation, we used general-purpose LLMs exclusively for grammar checking, phrasing refinement, and clarity improvements in the English text. All conceptual contributions, experiment designs, analyses, and claims in this paper are the responsibility of the authors. 

\section{Experimental Details}
\subsection{BenchMarks}\label{appendix:benchmark}
\textbf{BFCL} The Berkeley Function-Calling Leaderboard (BFCL-V3)~\citep{BFCL} constitutes a broad and systematic framework designed to rigorously evaluate the function-calling proficiency of large language models (LLMs) across a diverse spectrum of programming languages, application domains, and intricate real-world scenarios. The benchmark encompasses tasks ranging from multiple and parallel function invocations to multi-turn and multi-step function-call interactions. In total, BFCL-V3 comprises 4,951 test instances—3,951 single-turn cases and 1,000 multi-turn samples-carefully curated to reflect dynamic, authentic use cases. The assessment methodology in BFCL incorporates several complementary metrics: 
\begin{itemize}
    \item \textbf{Abstract Syntax Tree (AST) Evaluation}: This metric examines the structural correspondence between the abstract syntax tree of the model-generated output, the ground-truth reference, and the formal function specification. It evaluates the correctness of function identification, the inclusion and accuracy of obligatory parameters, and the precision of both parameter types and associated values.
    \item \textbf{Executable Function Evaluation}: Here, the produced API call is executed, and its runtime output is compared directly against the expected ground-truth result, thereby measuring practical execution accuracy.
    \item \textbf{Multi-turn State-based Evaluation}: The evaluation focus on comparing the backend system's state after all function calls are executed at the end of each turn of the conversation. It capture the correctness of model executions that modify the internal state via write and delete. 
    \item \textbf{Multi-turn Response-based Evaluation}: It compares the model's execution path against the minimial viable execution result paths as labeled in ground truth. The minimial viable execution result paths refer to a list of function calls that must be executed in order to produce desired response as user requests.
    \item \textbf{Irrelevance}: This criterion quantifies the model’s capacity to avoid generating function calls when presented with extraneous or unrelated user queries. The irrelevance score is determined by dividing the number of accurate non-function-call responses by the total test set size.
    \item \textbf{Relevance}: Relevance gauges the model’s adeptness at producing function calls that align contextually with the user’s query, irrespective of parameter value accuracy. This score is computed as the proportion of appropriate function-call responses within the entire evaluation set.
\end{itemize}

\textbf{ACEBench} ACEBench~\citep{ACEBench} is designed to evaluate tool-use capabilities with fine-grained categorization which could be divided into three primary categories: Normal, Special, Agent.``Normal" evaluates tool usage in basic scenarios;``Special" evaluates tool usage in situations with ambiguous or incomplete instructions;``Agent" evaluates tool usage through multi-agent interactions to simulate real-world, multi-turn dialogues:
\begin{itemize}
    \item \textbf{Normal Evaluation} compares the model’s function call output with the ground truth using AST parsing. 
    \item \textbf{Special Evaluation} mainly assesses the ability of model in problem identification. Specifically, the model must: (1) detect and alert missing parameters, (2) accurately locate erroneous parameters, and (3) recognize task-function mismatches. 
    \item \textbf{Agent Evaluation} focus on the model's proficiency in utilizing tools during human-agent interactions, employing gpt-4o as a user simulator, incluing End-to-End Accuracy and Process Accuracy. 
\end{itemize}

\subsection{Hyper-Parameters}\label{appendix:hyperparameter}
\begin{table}[h]
\centering
\caption{Configuration for Iterative GRPO training.}
\begin{tabular}{ll}
\hline
\textbf{Category} & \textbf{Hyperparameter} \\ \hline
\multirow{4}{*}{Data Configuration} & Train Batch Size: 128\\
 & Validation Batch Size: 128 \\
 & Max Prompt Length: 4096\\
 & Max Response Length: 1024 \\
\hline
\multirow{6}{*}{Optimization}& Learning Rate: 1e-6 \\
 & PPO Mini Batch Size: 128 \\
 & KL Loss Used: False \\
 &Entropy Loss Used: False\\ 
 &Clip Ratio Low: 0.2 \\
 &Clip Ratio High: 0.28 \\
\hline
\multirow{3}{*}{Rollout Configuration} & Rollout Mini Batch Size: 2\\
 & GPU Memory Utilization: 0.7\\
 & Number of Rollouts: 16\\
\hline
\multirow{2}{*}{Training \& Logging} & Save Frequency (epoch): 1\\
 & Test Frequency (epoch): 1\\ \hline
\end{tabular}
\end{table}

\section{The Algorithm of LoopTool}
We present the complete procedure of our \textbf{LoopTool} framework in Algorithm~\ref{Algorithm:LoopTool} , which couples \emph{GRPO-based post-training}, \emph{Greedy Capability Probing (GCP)}, \emph{Judgement-Guided Label Verification (JGLV)}, and \emph{Error-Driven Data Expansion (EDDE)} into a unified closed-loop data evolution process.

\begin{algorithm}[!ht]
\caption{LoopTool: Iterative Model-Aware Data Evolution Framework}
\label{Algorithm:LoopTool}
\KwIn{
Initial seed dataset $\mathcal{D}_{\text{seed}}$ from Automated Tool-Augmented Data Construction;
Initial model parameters $\pi_{\theta_0}$.
}
\KwOut{
Final optimized tool-calling model $\pi_{\theta_J}$ after $J$ iterations.
}

\BlankLine
\textbf{Initialize:} $j \gets 1$, $\mathcal{D}_1 \gets \text{Subset}(\mathcal{D}_{\text{seed}})$.

\While{$j \leq J$}{
    \tcp{Step 1: GRPO-based Post-training}
    Train policy $\pi_{\theta_{j-1}}$ on $\mathcal{D}_j$ using GRPO in Eq.(\ref{eq:GRPO}) with binary reward $r(\cdot)$, obtaining updated parameters $\pi_{\theta_j}$.
    
    \BlankLine
    
    \tcp{Step 2: Greedy Capability Probing (GCP)}
    \ForEach{$(\mathcal{T}, c_t, a_t^*) \in \mathcal{D}_j$}{
        Generate $a_t$ via deterministic greedy decoding from $\pi_{\theta_j}$\;
        \If{$a_t \neq a_t^*$}{
            Send $(\mathcal{T}, c_t, a_t^*; a_t)$ to JGLV for evaluation\;
        }
        Compute $\mathrm{PPL}_{(\mathcal{T}, c_t)}$ by Eq.(\ref{eq:PPL}) and retain high-PPL samples and $a_t=a_t^*$ into $\mathcal{D}_j^{HPPL}$\;
    }
    
    \BlankLine
    
    \tcp{Step 3: Judgement-Guided Label Verification (JGLV)}
    \ForEach{mismatched case $(\mathcal{T}, c_t, a_t^*; a_t)$}{
        Obtain judgement result $y_{\mathrm{judge}} \in \{\texttt{PRED\_WRONG}, \texttt{LABEL\_WRONG}, \texttt{BOTH\_CORRECT}, \texttt{BOTH\_WRONG}\}$ via Qwen3-32B\;
        \uIf{$y_{\mathrm{judge}} = \texttt{PRED\_WRONG}$}{
            Add to $\mathcal{D}_j^{MR}$\;
        }
        \uElseIf{$y_{\mathrm{judge}} = \texttt{LABEL\_WRONG}$}{
            Replace $a_t^* \gets a_t$ and add to $\mathcal{D}_j^{LR}$\;
        }
        \uElseIf{$y_{\mathrm{judge}} \in \{\texttt{BOTH\_CORRECT}, \texttt{BOTH\_WRONG}\}$}{
            Discard sample\;
        }
    }
    
    \BlankLine
    
    \tcp{Step 4: Error-Driven Data Expansion (EDDE)}
    Construct error seed set $\mathcal{D}_j^{ES} \gets \mathcal{D}_j^{MR} \cup \mathcal{D}_j^{LR}$\;
    \ForEach{error seed in $\mathcal{D}_j^{ES}$}{
        Generate $k$ new samples with scenario diversification constraints\;
    }
    Validate generated set via rule-based + LLM-based evaluation to obtain $\mathcal{D}_j^{EE}$\;
    
    \BlankLine
    
    \tcp{Step 5: Dataset Update for Next Iteration}
    Select untrained subset $\mathcal{D}_j^{\text{Seed-new}} \subset \mathcal{D}_{\text{seed}}$\;
    Construct next-round dataset by Eq.(\ref{eq:next_iter_dataset}):
    \[
        \mathcal{D}_{j+1} = \mathcal{D}_j^{ES} \cup \mathcal{D}_j^{EE} \cup \mathcal{D}_j^{HPPL} \cup \mathcal{D}_j^{\text{Seed-new}}
    \]
    
    \BlankLine
    $j \gets j+1$\;
}

\Return{$\pi_{\theta_J}$}
\end{algorithm}

\section{The example of Hierarchical Dual SubTrees}\label{appendix:dual_tree_examples}
\begin{figure}[ht]
    \centering
    \begin{minipage}[t]{0.48\linewidth}
        \centering
        \includegraphics[width=\linewidth]{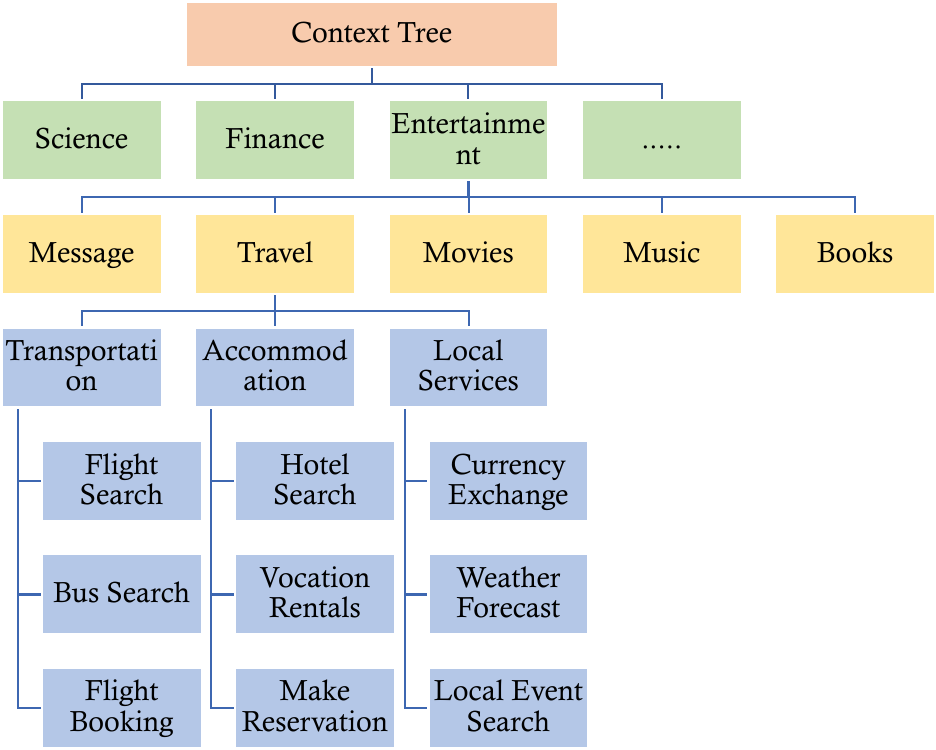}
        \caption{The example subtree of Context Tree.}
        \label{Fig:Context_Tree}
    \end{minipage}
    \hfill
    \begin{minipage}[t]{0.48\linewidth}
        \centering
\includegraphics[width=\linewidth]{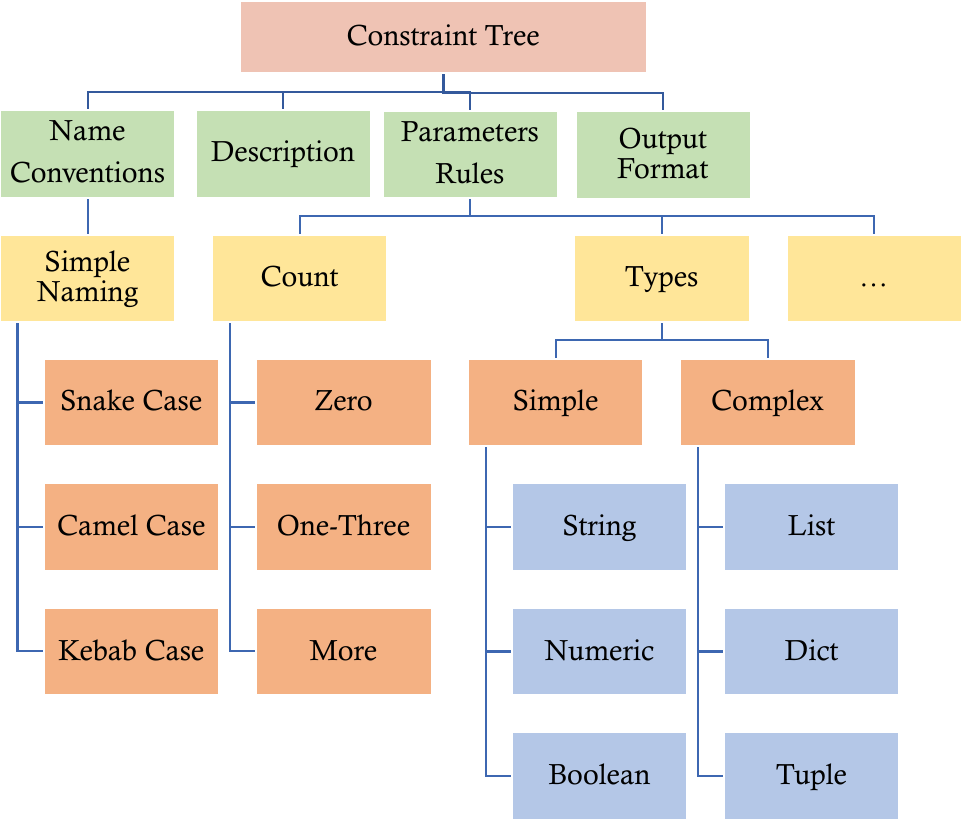}
        \caption{The example subtree of Constraint Tree.}
    \label{Fig:Constraint_Tree}
    \end{minipage}
\end{figure}

The example subtrees of the Context Tree and Constraint Tree are illustrated in Figure~\ref{Fig:Context_Tree} and Figure~\ref{Fig:Constraint_Tree}, respectively.

\section{The Training Sample for GRPO}\label{appendix:train_format}

\begin{figure}[htbp]
    \centering
    \includegraphics[width=1.0\linewidth]{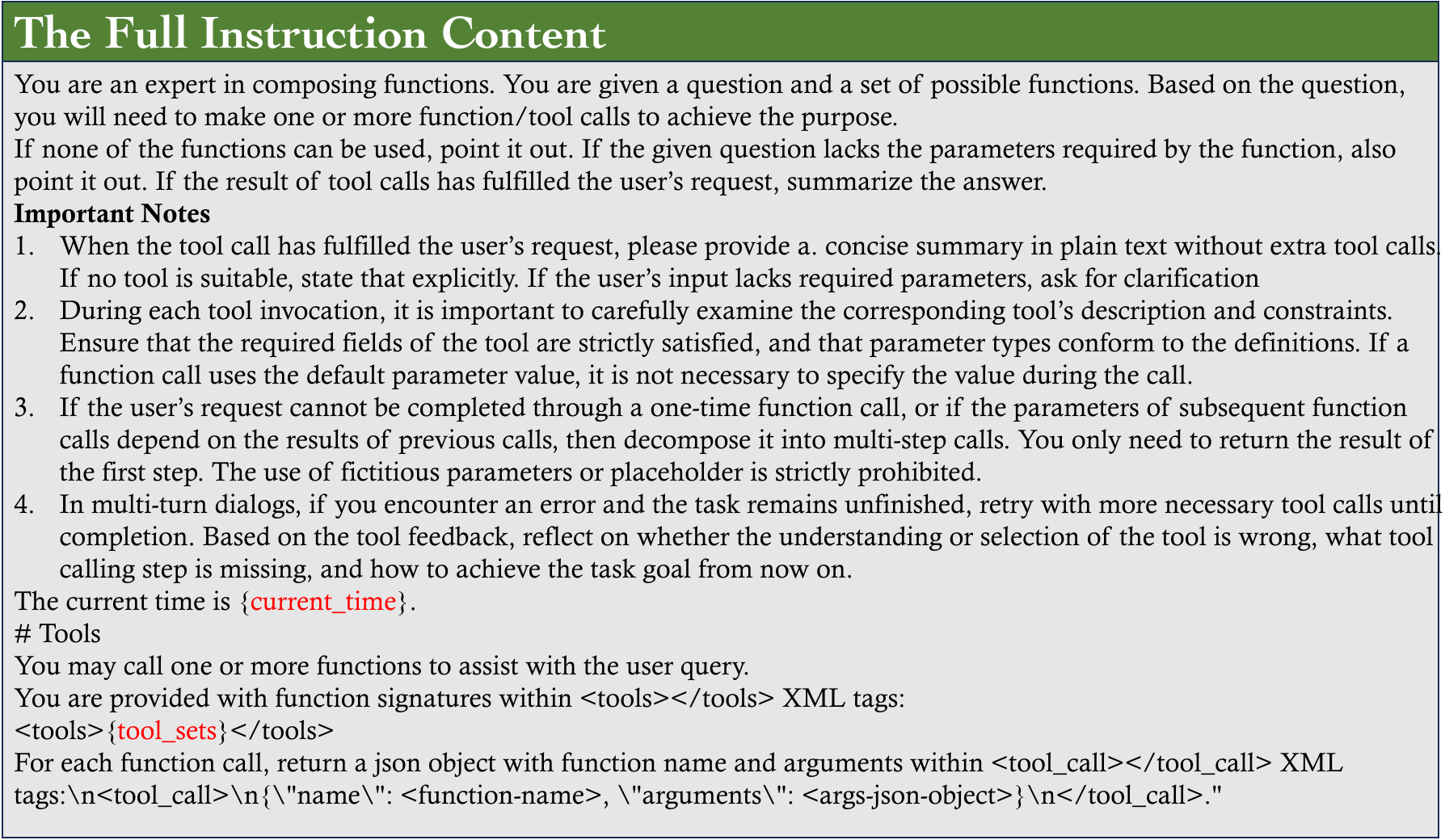}
    \captionsetup{skip=1pt} 
    \caption{The general instruction prompt employed in all GRPO samples. The variables $current\_time$ and $tool\_sets$ are placeholders.}
    \label{Fig:Instruction_template}
\end{figure}

\begin{figure}[htbp]
    \centering
    \includegraphics[width=1.0\linewidth]{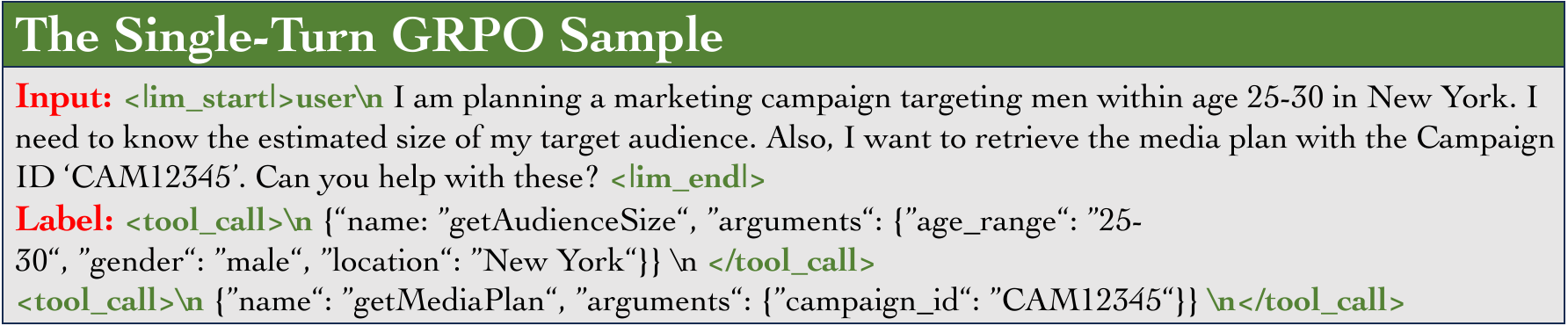}
    \captionsetup{skip=1pt} 
    \caption{The example of Single-Turn GRPO samples.}
    \label{Fig:single_turn_sample}
\end{figure}

\begin{figure}[htbp]
    \centering
    \includegraphics[width=1.0\linewidth]{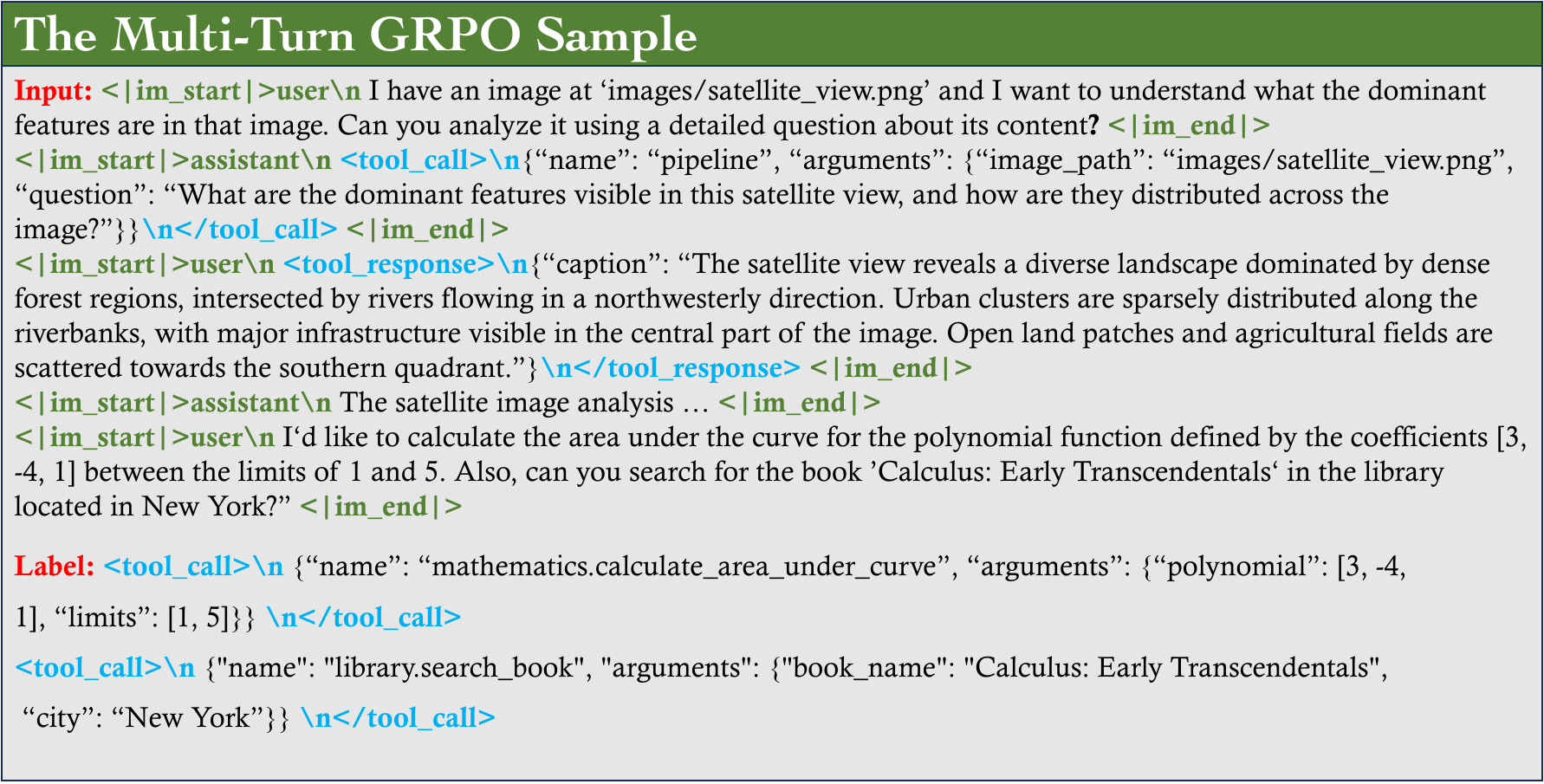}
    \captionsetup{skip=1pt} 
    \caption{The example of Multi-Turn GRPO samples.}
    \label{Fig:multi_turn_sample}
\end{figure}

The Instruction Prompt used in all GRPO samples is illustrated in Figure~\ref{Fig:Instruction_template}. The Single-Turn and Multi-Turn samples are illustrated in Figure~\ref{Fig:single_turn_sample} and Figure~\ref{Fig:multi_turn_sample}.

\section{The Label Verification Prompt}\label{appendix:JGLV}
\begin{figure}[htbp]
    \centering
\includegraphics[width=1.0\linewidth]{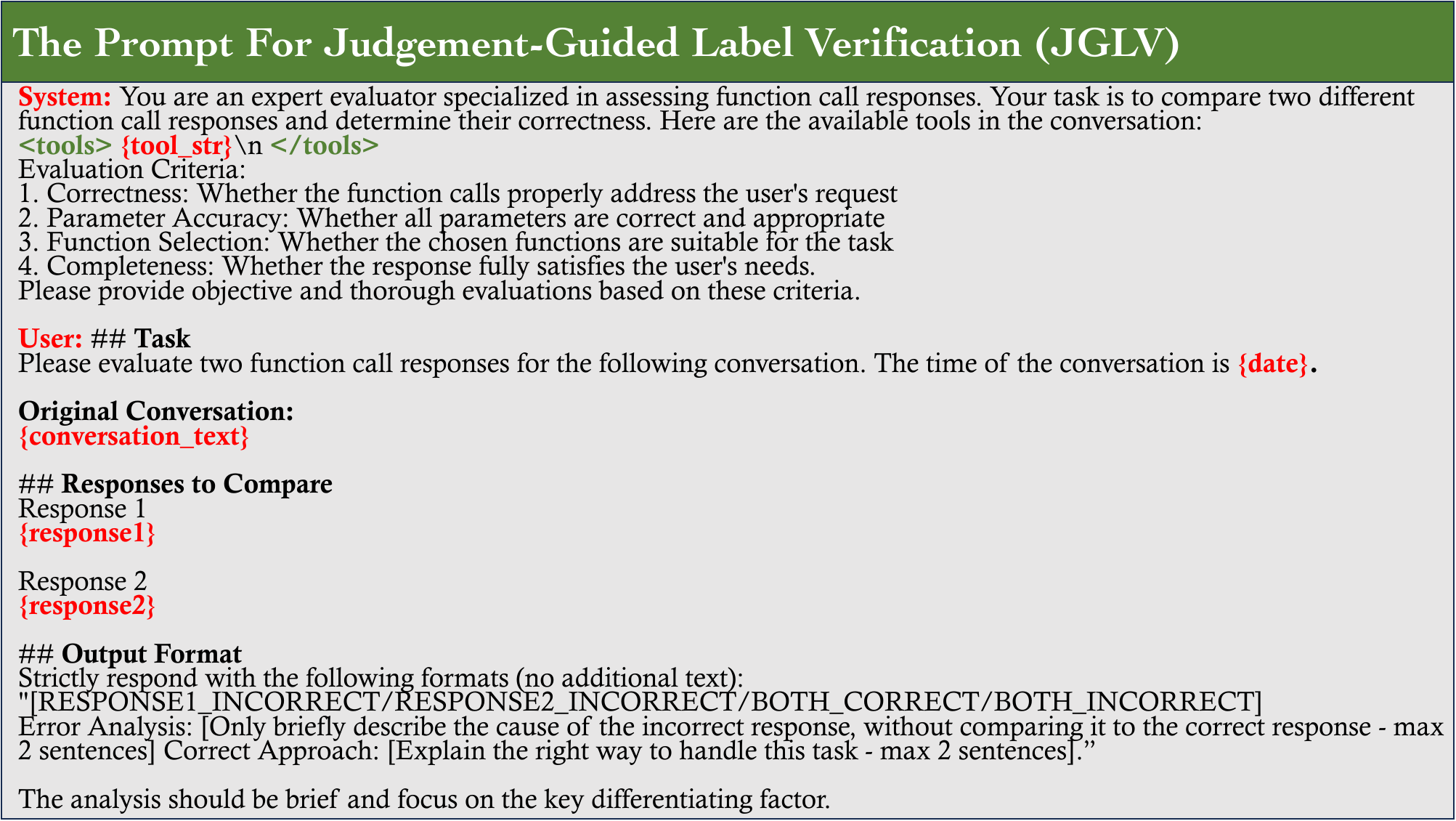}
    \captionsetup{skip=2pt} 
    \caption{The Prompt used in Judge-Guide Label Verification for Judgement Model. The red text corresponds to variables that are placeholders.}
    \label{Fig:JGLV_Prompt}
\end{figure}

\begin{figure}[htbp]
    \centering
\includegraphics[width=1.0\linewidth]{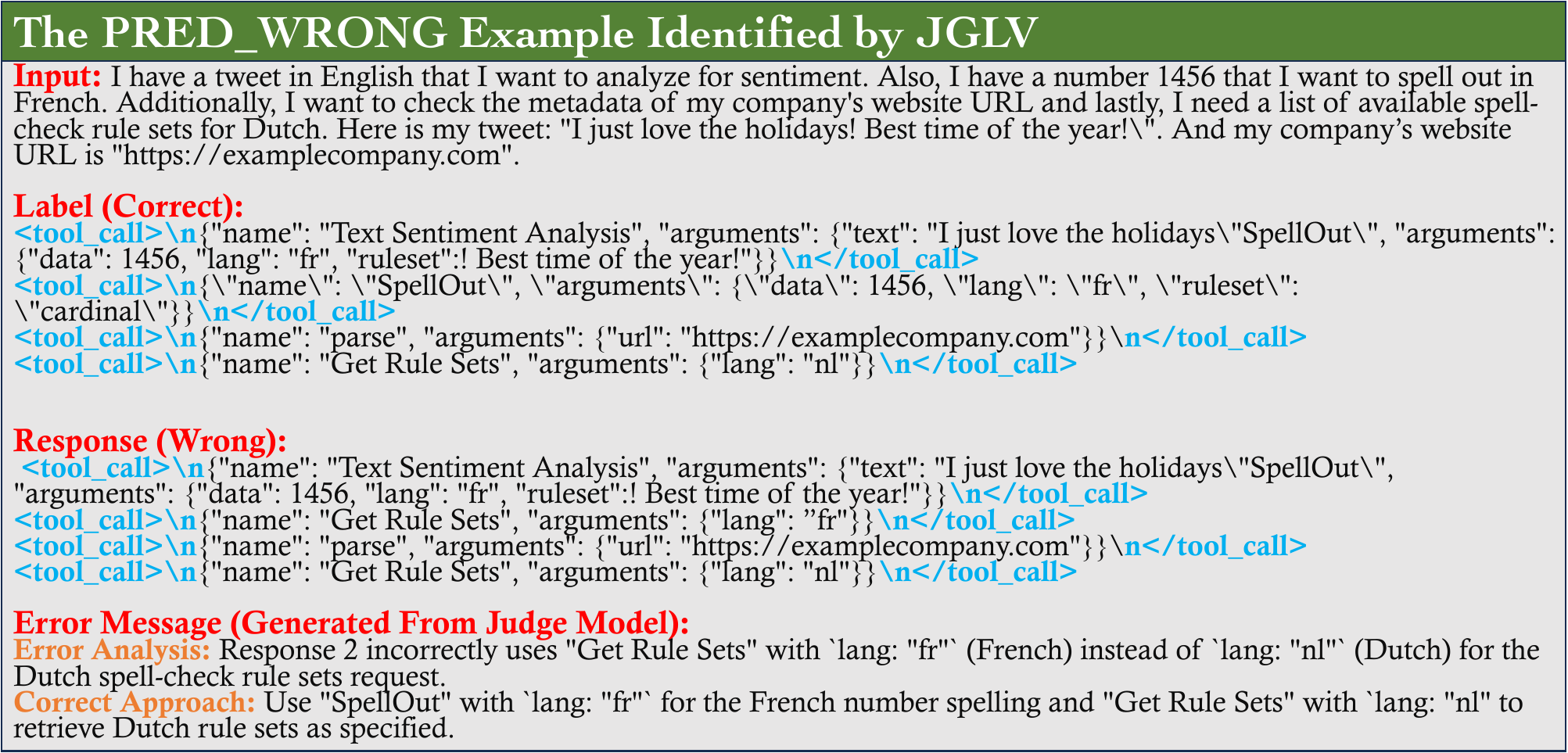}
    \caption{The example with $y_{judge} = \texttt{PRED\_WRONG}$ identified by JGLV.}
    \label{Fig:PRED_WRONG}
\end{figure}

\begin{figure}[htbp]
    \centering
\includegraphics[width=1.0\linewidth]{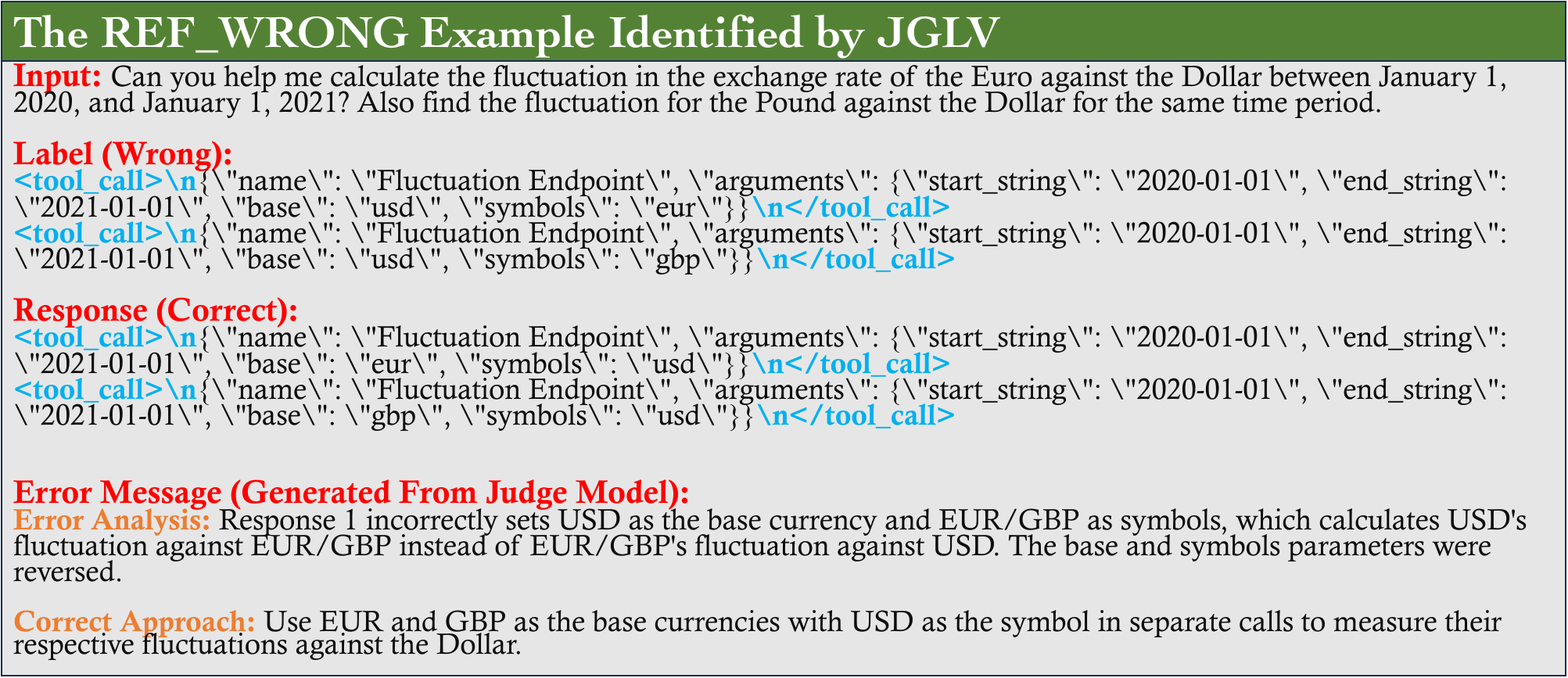}
    \caption{The example with $y_{judge} = \texttt{REF\_WRONG}$ identified by JGLV.}
    \label{Fig:REF_WRONG}
\end{figure}

The Prompt used in Judge-Guide Label Verification (JGLV) is concluded in Figure~\ref{Fig:JGLV_Prompt}. Sample examples with $y_{judge} = \texttt{PRED\_WRONG}$ and $y_{judge} = \texttt{REF\_WRONG}$ are respectively presented in Figures~\ref{Fig:PRED_WRONG} and~\ref{Fig:REF_WRONG}.

\section{The Error Generation Prompt and New Error Samples}\label{appendix:EDDE}
\begin{figure}[htbp]
    \centering
\includegraphics[width=1.0\linewidth]{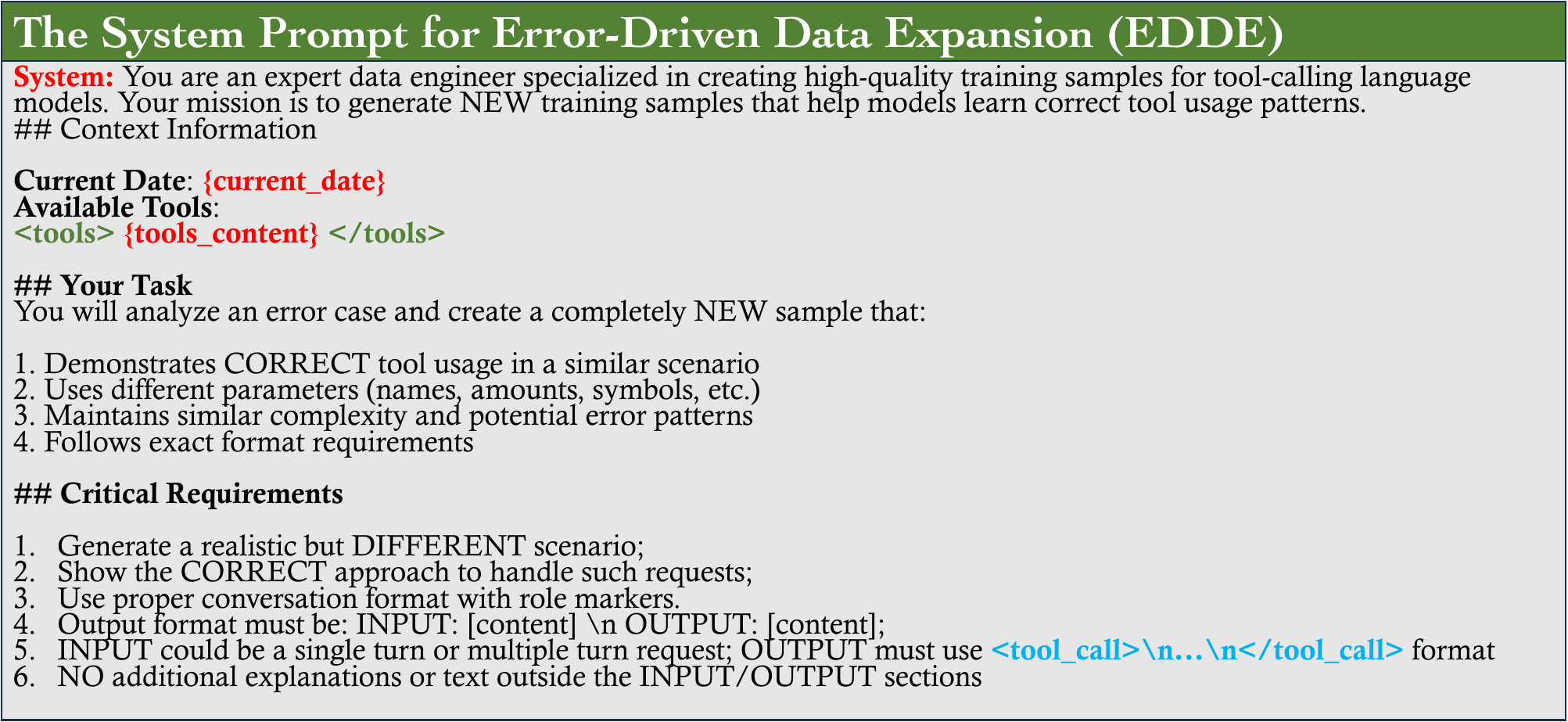}
    \caption{The system prompt for Error-Driven Data Expansion (EDDE). }
    \label{Fig:System_Prompt_EDDE}
\end{figure}

\begin{figure}[htbp]
    \centering
\includegraphics[width=1.0\linewidth]{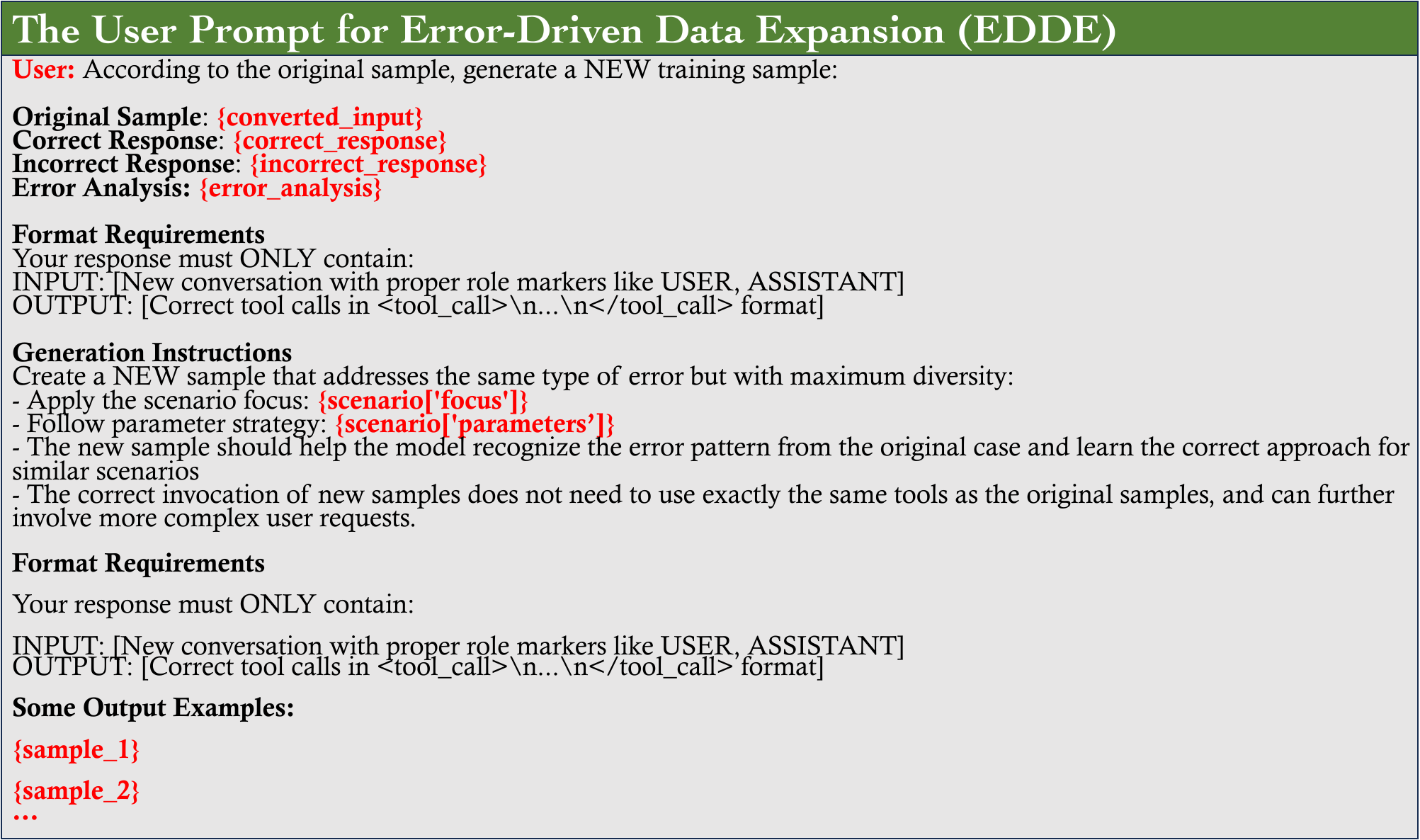}
    \caption{The user prompt for Error-Driven Data Expansion (EDDE).}
    \label{Fig:User_Prompt_EDDE}
\end{figure}

\begin{figure}[htbp]
    \centering
\includegraphics[width=1.0\linewidth]{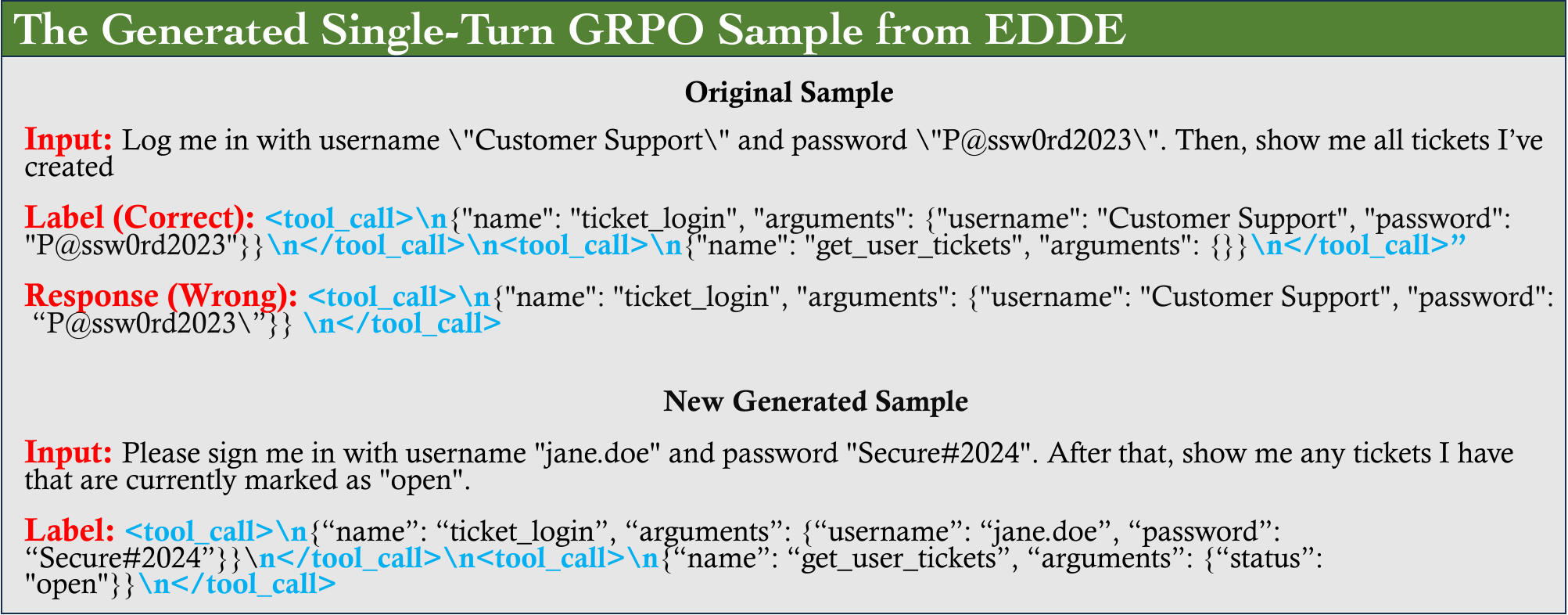}
    \caption{The new sample generated by EDDE according to the error in the model response.}
    \label{Fig:New_Sample}
\end{figure}

The system and user prompts for Error‑Driven Data Expansion (EDDE) are illustrated in Figures~\ref{Fig:System_Prompt_EDDE} and~\ref{Fig:User_Prompt_EDDE}, respecitively. The generated sample case is shown in Figure~\ref{Fig:New_Sample}.

\section{The Learning Curves in Iterative Learning}\label{appendix:learning_curve}

\end{document}